%% file: causalitylaw.tex
\documentclass[12pt]{article}
\usepackage{chicagob}
\usepackage{times}
  \usepackage{graphicx}

\input{defn.tex}

\renewcommand{\S}{{\cal S}}
\newcommand{\Suff}{\mathbf{S}}
\newcommand{\ML}{\mathit{MD}}

\renewcommand{\FF}{\mathit{FF}}

\newcommand{\HU}{\mathit{HU}}
\newcommand{\HR}{\mathit{HR}}
\newcommand{\NW}{\mathit{NW}}
\newcommand{\VS}{\mathit{VS}}

\newcommand{\WB}{\mathit{WB}}
\newcommand{\BMC}{\mathit{BMC}}
\newcommand{\TT}{\mathit{TT}}

\newcommand{\ST}{\mathit{ST}}

\newcommand{\MT}{\mathit{MT}}
\newcommand{\CP}{\mathit{CP}}
\newcommand{\AC}{\mathit{DAP}}
\newcommand{\PD}{\mathit{PD}}

\newcommand{\BH}{\mathit{BH}}
\renewcommand{\BS}{\mathit{BS}}
\newcommand{\BT}{\mathit{BT}}
\newcommand{\SH}{\mathit{SH}}

\newcommand{\dr}{\mbox{{\em dr}}}
\newcommand{\db}{\mbox{{\em db}}}
\newcommand{\WIN}{\mbox{{\em WIN}}}

\begin{document}

\begin{titlepage}
\title{Cause, Responsibility, and Blame: \\
A Structural-Model Approach}
\author{ }
\author{Joseph Y. Halpern%
\thanks{Supported in part by NSF under
under grants ITR-0325453, IIS-0534064, and IIS-0911036,
and by AFOSR under grant 
FA9550-05-1-0055.}\\
Cornell University\\
Dept. of Computer Science\\
Ithaca, NY 14853\\
halpern@cs.cornell.edu\\
http://www.cs.cornell.edu/home/halpern}

\setcounter{page}{0}
\thispagestyle{empty}
\maketitle
\thispagestyle{empty}

\begin{abstract}
A definition of causality introduced by Halpern
and Pearl \citeyear{HP01b}, which uses \emph{structural equations},
is reviewed.  A more refined definition is
then considered, which takes into account issues of normality and
typicality, which are well known to affect causal ascriptions.
Causality is typically an all-or-nothing notion: either $A$ is a cause
of $B$ or it is not.  An extension of the definition of causality to
capture notions of \emph{degree of
responsibility} and \emph{degree of
blame}, due to Chockler and Halpern \citeyear{ChocklerH03}, is reviewed.
For example, if someone wins an  election 11--0, then each person who votes 
for him is less responsible for the victory than if he had won 6--5, and
another extension that considers the \emph{degree of blame}, which takes
into account an agent's epistemic state.  Roughly speaking, the degree
of blame of $A$ for $B$ is the expected degree of responsibility of $A$
for $B$, taken over the epistemic state of an agent.  Finally, the
structural-equations definition of causality is compared to Wright's
\citeyear{Wright85,wright:88,Wright01} NESS test.
\end{abstract}
\end{titlepage}

\begin{center}
{\LARGE {\bf Cause, Responsibility, and Blame: \\A Structural-Model Approach}}
\end{center}

\section{Introduction}
It is generally agreed that the notion of legal cause is sorely in need
of clarification.  Not surprisingly, there has been a great deal of
effort in the legal community to provide that clarification (see
\cite{HH85,Wright85,wright:88} and the references therein).
Philosophers have also spent a great deal of effort attempting to
clarify causality (see \cite{Collins03} for a recent collection of
articles on the subject, along with pointers to the literature).  
Recently there has also been work on causality be computer scientists.
It is that work that I report on here.  In particular, I describe a 
definition of causality due to Halpern and Pearl \citeyear{HP01b};
I henceforth call this the HP definition.

The HP definition is more formal and mathematical than other
definitions of causality in the literature. 
While this does add some initial overhead, it has the advantage of there
being far less ambiguity.   There is no need, as
in most other definitions, to try to understand how to interpret the
English (what counts as a ``sufficient condition''? what is a ``part''
of a sufficient condition?).   The first step in the HP definition involves
building a formal model in which causality can be
determined unambiguously.  The definition will then say only that $A$
is a cause of $B$ \emph{in model $M$}.  It is possible 
to construct two closely related models models $M_1$ and $M_2$ such that
$A$ is a cause of $B$ in $M_1$ but not in $M_2$.  There is not
necessarily a ``right'' model (and, in any case, the definition is
silent on what makes one model better than another, although see
\cite{HH10} for some discussion of this issue).  That means that,
even if there is agreement regarding the definition of causality, there
may be disagreement about which model better describes
the real world.  I view this as a feature of the
definition.  It moves the question of actual causality to the right
arena---debating which of two (or more) models of the world is a better
representation of those aspects of the world that one wishes to capture
and reason about.  This, indeed, is the type of debate that goes on in
informal (and legal) arguments all the time.

The HP definition has several other advantages.  For one thing, it can
be used to provide a definition of \emph{explanation} \cite{HP01a}, an
issue I do not discuss in this paper.  For another, it can accommodate
reasoning about normality and typicality, which is well known to affect
causal ascriptions \cite{KM86}.  A third advantage is that, as 
Chockler and Halpern \citeyear{ChocklerH03} showed,  it can be extended 
to capture notions of \emph{degree of responsibility} and \emph{degree of
blame}.  To understand how this is done, first note that causality is an 
all-or-nothing concept.  That is, $A$ is either a cause of $B$ or it is
not.  As a consequence, thinking only in terms of causality prevents
us from making what may be important distinctions.  For
example, suppose that Mr.~B wins an election against Mr.~G by a vote
of 11--0. According to the HP definition, each of the people who voted
for Mr.~B is a cause of him winning. However, it seems that their degree
of responsibility should not be as great as in the case when Mr.~B wins
6--5.  Chockler and Halpern suggested that the degree of responsibility of $A$
for $B$ should be taken to be $1/(N+1)$, 
where $N$ is the minimal number of changes that have to be made to
obtain a situation where $B$ counterfactually depends directly on $A$,
one where, if $A$ hadn't happened, then $B$ wouldn't have happened.  
In the case of the 11--0 vote, each voter has degree of responsibility
$1/6$ for Mr.~B's victory, since the votes of five other voters have to
be change (giving a vote of 6--5) before Mr.~B's victory
counterfactually depends directly on the vote of any of the six
remaining people who voted for him; if any of these six voters had
changed their vote, then Mr.~G would have won.  By way of contrast, 
in the case of a 6--5 vote, each of the six voters is already directly
counterfactually responsible for Mr.~B's victory; if any of them had
changed their vote, then Mr.~B would not have won.  Thus, each has
degree of responsibility 1.  
In general, the degree of responsibility of $A$ for $B$ is between 0 and
1.  It is 0 if and only if $A$ is not a cause of $B$; it is 1 if $B$
directly counterfactually depends on $A$; it is strictly between 0 and 1
otherwise.  

This definition of degree of responsibility implicitly assumes that 
the facts of the world and how the world works are known.  For example,
when saying that voter 1 has degree of responsibility $1/6$ for Mr.~B's
win when the vote is 11--0, we assume that the vote and the
procedure for determining a winner (majority wins) is known.  There is
no uncertainty about this.  But this misses out on important
component of determining what Chockler and Halpern called {\em blame}:~the epistemic
state.  Consider a doctor who treats a patient with a particular drug
resulting in the patient's death.  The doctor's treatment is a cause of
the patient's death; indeed, the doctor may well bear degree of
responsibility 1 for the death.  However, if the doctor had no idea that
the treatment had adverse side effects for people with  high
blood pressure, he should perhaps not be held to blame for the death.
Actually, in legal arguments, it may not be so relevant what the doctor
actually did or did not know, but what he {\em should have known}.
Thus, rather than considering the doctor's actual epistemic state, it
may be more important to consider what his epistemic state should have
been.  Chocker and Halpern \citeyear{ChocklerH03} give a definition of
\emph{degree of blame} that takes this into account; roughly speaking,
the degree of blame of an agent who performed $A$ has for $B$ is the
expected degree of responsibility of $A$ for $B$, relative to agent's
epistemic state.  

To understand the difference between responsibility and blame, suppose
that there is a firing squad consisting of ten excellent marksmen.  Only
one of them has live bullets in his rifle; the rest have blanks.  
The marksmen do not know which of them has the live bullets.  The
marksmen shoot at the prisoner and he dies.  The only marksman that is
the cause of the prisoner's death is the one with the live bullets.
That marksman has degree of responsibility 1 for the death; all the rest
have degree of responsibility 0.  However, each of the marksmen has
degree of blame $1/10$.

The rest of this paper is organized as follows.  
In Section~\ref{sec:causalmodel}, I provide a gentle introduction to
structural equations and causal models.  In Section~\ref{sec:actcaus}, I
review the HP definition of actual causality.  In
Section~\ref{sec:final}, I show how the definition can be refined to
deal with normality and typicality.  In
Section~\ref{sec:responsibility}, I show how the definition can be
extended to capture responsibility and blame.  In
Section~\ref{sec:NESS}, I compare the
structural-model definition of causality is compared to Wright's
\citeyear{Wright85,wright:88,Wright01} NESS test,
which is perhaps the closest work in the
legal literature to it.    I conclude in
Section~\ref{sec:conc}.  
Sections~\ref{sec:causalmodel} and \ref{sec:actcaus} are largely taken 
from \cite{HP01b}; Section~\ref{sec:final} is based on material from
\cite{HH11}; Section~\ref{sec:responsibility} is based on material
from \cite{ChocklerH03}; and material in Section~\ref{sec:NESS}
appeared in preliminary form in \cite{Hal39}.   The reader is
encouraged to 
consult these papers for more details, more intuition, and more
examples.  


\section{Causal Models}\label{sec:causalmodel}

In this section, I review the formal model of causality used in the HP
definition.  The description of causal models given here is taken
from \cite{Hal20}; it is a formalization of earlier work of Pearl
\citeyear{Pearl.Biometrika}, further developed in 
\cite{GallesPearl97,Hal20,pearl:2k}.

The model assumes that the world is described in terms of \emph{random
variables} and their values.  For example, if we are trying to determine
whether a forest fire was caused by lightning or an arsonist, we 
can take the world to be described by three random variables:
\begin{itemize}
\item $\FF$ for forest fire, where $\FF=1$ if there is a forest fire and
$\FF=0$ otherwise; 
\item $L$ for lightning, where $L=1$ if lightning occurred and $L=0$ otherwise;
\item $\ML$ for match dropped (by arsonist), where $\ML=1$ if the arsonist
dropped a lit match, and $\ML = 0$ otherwise.
\end{itemize}
Similarly, in the case of the 11--0 vote, we can describe the world by
12 random variables, $V_1, \ldots, V_{11}, W$, where $V_i = 0$ if voter
$i$ voted for Mr.~B and $V_1 = 1$ if voter $i$ voted for Mr.~G, for $i =
1, \ldots, 11$, $W = 0$ if Mr.~B wins, and $W=1$ if Mr.~G wins.  

In these two examples, all the random variables are \emph{binary}, that
is, they take on only two values.  There is no problem allowing a random
variable to have more than two possible values.  For example, the random
variable $V_i$ could be either 0, 1, or 2, where $V_i= 2$ if $i$ does
not vote; similarly, we could take $W=2$ if the vote is tied, so neither
candidate wins.

The choice of random variables determines the language used to frame
the situation.  Although there is no ``right'' choice, clearly some
choices are more appropriate than others.  For example, when trying to
determine the cause of Sam's lung cancer, if there is no random variable
corresponding to smoking in a model then, in that model, we will not
be able to conclude that smoking is a cause of Sam's lung cancer.  

Some random variables may have a causal influence on others. This
influence is modeled by a set of {\em structural equations}.
For example, if we want to model the fact that
if the arsonist drops a match \emph{or} lightning strikes then a fire
starts, we could use 
the random variables $\ML$, $\FF$, and $L$ as above, with the equation
$\FF = \max(L,\ML)$; that is, the value of the random variable $\FF$ is the
maximum of the values of the random variables $\ML$ and $L$.  This 
equation says, among other things, that if $\ML=0$ and $L=1$, then $\FF=1$.
The equality sign in this equation should be thought of more like an 
assignment statement in programming languages; once we set the values of
$\FF$ and $L$, then the value of $\FF$ is set to their maximum.  However,
despite the equality, if a forest fire starts some other way, that does not
force the value of either $\ML$ or $L$ to be 1.

Alternatively, if we want to model the fact that a fire requires both a
lightning strike \emph{and} a dropped match (perhaps the wood is so wet
that it needs two sources of fire to get going), then the only change in the
model is that the equation for $\FF$ becomes $\FF = \min(L,\ML)$; the
value of $\FF$ is the minimum of the values of $\ML$ and $L$.  The only
way that $\FF = 1$ is if both $L=1$ and $\ML=1$.  

Both of these models are somewhat simplistic.  Lightning does not always
result 
in a fire, nor does dropping a lit match.  One way of dealing with this
would be to make the assignment statements probabilistic.  For example,
we could say that the probability that $\FF=1$ conditional on $L=1$
is .8.  This approach would lead to rather complicated definitions.
It is much simpler to think of all the equations as being deterministic
and, intuitively, use enough variables to capture all the conditions
that determine whether there is a forest fire are captured by
random variables.  One way to do this is simply to add those variables
explicitly.  For example, we could add variables that talk about the
dryness of the wood, the amount of undergrowth, the presence of
sufficient oxygen  (fires do not start so easily on the top of high
mountains), and so on.  If a modeler does not want to add all these
variables explicitly (the details may simply not be relevant to the
analysis), another alternative is to use a single variable,
say $U$, which intuitively incorporates all the relevant factors,
without describing them explicitly.  The value of $U$ would determine
whether the lightning strikes, whether the match is dropped by
the arsonist, and whether both are needed to start a fire, just one, or
neither (perhaps fires start spontaneously, or there is a cause not
modeled by the random variables used).  In this way of modeling things,
$U$ would take on 8 possible values of the form $(i,j,k)$, where $i$,
$j$, and $k$ are each either 0 or 1.  Intuitively,
$i$ describes whether the external conditions are such that the
lightning strikes (and encapsulates all the conditions, such as humidity
and temperature, that 
affect whether the lightning strikes); $j$ describes whether 
the arsonist drops the match (and thus encapsulates the psychological
conditions that determine whether the arsonist drops the match); and $k$
describes whether just one of or both a dropped match and
lightning are needed for the forest fire.
Thus, the equation could say that $\FF=1$ if, for example,
$U=(1,1,1)$ (so that the lightning strikes, the match is dropped, and
only one of them is needed to have a fire) or if $U = (1,1,2)$, 
but not if $U = (1,0,2)$.  For future reference, let
$U_1$, $U_2$, and $U_3$ denote the three components of the value of $U$
in this example, so that if $U=(i,j,k)$, then $U_1 = i$, $U_2 =j$, and
$U_3 = k$.

It is conceptually useful to split the random variables into two
sets, the {\em exogenous\/} variables, whose values are determined by
factors outside the model, and the
{\em endogenous\/} variables, whose values are ultimately determined by
the exogenous variables.  In the forest-fire example, the
variable $U$ is exogenous, while the variables $\FF$, $L$, and $\ML$ are
endogenous. We want to take as given that there is 
enough oxygen for the fire, that the wood is sufficiently dry to burn,
and whatever factors make the arsonist drop the match.  Thus, it is only
the endogenous variables whose values are described by the
structural equations.  

With this background, we can formally define a \emph{causal model} $M$
as a pair $(\S,\F)$, where $\S$ is a \emph{signature}, which explicitly
lists the endogenous and exogenous variables  and characterizes
their possible values, and $\F$ defines a set of \emph{modifiable
structural equations}, relating the values of the variables.  In the
next two paragraphs, I  define $\S$ and $\F$ formally; the definitions
can be skipped by the less mathematically inclined reader.  

A signature $\S$ is a tuple $(\U,\V,\R)$, where $\U$ is a set of
exogenous variables, $\V$ is a set 
of endogenous variables, and $\R$ associates with every variable $Y \in 
\U \union \V$ a nonempty set $\R(Y)$ of possible values for $Y$
(i.e., the set of values over which $Y$ {\em ranges}).  As suggested
above, in the
forest-fire example, we have $\U = \{U\}$, 
$\R(U)$ consists of the 8 possible values of $U$ 
discussed earlier, $\V = 
\{\FF,L,\ML\}$, and $\R(\FF) = \R(L) = \R(\ML) = \{0,1\}$.  

$\F$ associates with each endogenous variable $X \in \V$ a
function denoted $F_X$ such that $F_X$ maps $\times_
{Z \in (\U \union
\V - \{X\})} \R(Z)$  to $\R(X)$.%
\footnote{Recall that $\U \union \V - \{X\}$ is the set consisting of all the
variables in either $\U$ or $\V$ that are not in $\{X\}$.}
This mathematical notation just makes precise the fact that 
$F_X$ determines the value of $X$,
given the values of all the other variables in $\U \union \V$.
In the running forest-fire example,
we would have $F_{\FF}(L,\ML,U) = 1$ if $U_3 = 1$ and 
$\max(L,\ML) = 1$ (so that at least one of $L$ or $\ML$ is 1);
similarly,  $F_{\FF}(L,\ML,U) = 1$ if $U_3 = 2$ and 
$\min(L,\ML) = 1$ (so that both $L$ and $\ML$ are 1).
Note that the
value of $F_{\FF}$ is independent of the first two components of $U$;
all that matters is $U_3$ and the values of $L$ and
$\ML$.  The first two components of $U$ are used in the equations for $L$
and $\ML$: $U_1$ determines the value of $L$ and $U_2$ determines
the value of $\ML$; specifically, $F_L(\ML,\FF,U) = U_1$ and 
$F_{\ML}(L,\FF,U) = U_2$.  Note that the values of $L$ and $\ML$ are
independent of the value of $\FF$.  

I sometimes simplify notation and write $X = Y+U$ instead of 
$F_X(Y,Y',U) = Y+U$.  With this simplified notation, the equations for
the forest-fire example become
$$\begin{array}{cll}
L &= &U_1\\
\ML &= &U_2\\
\FF &= &\left\{\begin{array}{ll}
\max(L,\ML) &\mbox{if $U_3 = 1$}\\
\min(L,\ML)  &\mbox{if $U_3 = 2$.}
\end{array} \right.
\end{array}
$$

Although I may write something like $X = Y+U$, I stress that 
the fact that $X$ is assigned  $Y+U$ does \emph{not} imply
that $Y$ is assigned $X-U$; that is, $F_Y(X,Z,U) = X-U$ does not
necessarily hold.
With this equation, if $Y = 3$ and $U = 2$, then
$X=5$, regardless of how $Y'$ is set.  
It is this asymmetry in the treatment of of the variables on the left-
and right-hand side of the equality sign that arguably leads to the
asymmetry of causality: if $A$ is a cause of $B$, it will not be the
case that $B$ is a cause of $A$.

The key role of the structural equations is to define what happens in the
presence of external \emph{interventions}.  For example, they describe what
happens if the arsonist does \emph{not} drop the match.
In this case, there is a forest fire exactly if there is lightning and
lightning by itself suffices for a fire; that is, if the exogenous
variable $U$ has the form $(i,1,1)$.  Understanding the effect of
interventions will be critical in defining causality.

Setting the value of some variable $X$ to $x$ in a causal
model $M = (\S,\F)$ results in a new causal model denoted $M_{X
\gets x}$.  In the new causal model, the equation for $X$ is very
simple: $X$ is just set to $x$; the remaining equations are unchanged.
More formally, 
$M_{X \gets x} = (\S,
\F_{X \gets x})$, where 
$\F_{X \gets x}$ is the result of replacing the equation for $X$ in $\F$
by $X = x$.
Thus, if $M$ is the causal model for the forest-fire example, then in
$M_{\ML \gets 0}$, the model where the arsonist does not drop the match,
the equation $\ML = U_2$ is replaced by $\ML = 0$.

\emph{Counterfactuals} (statements counter to fact) have often been
considered when trying to define causality.  These (causal) equations
can be given a straightforward counterfactual interpretation. 
An equation such as $x = F_X(u,y)$ should be thought of as saying 
that in a context where the exogenous variable $U$ has value $u$,
if $Y$ were set to $y$ by some means (not specified in the model), 
then $X$ would take on the value $x$, as dictated by $F_X$.
However, if the value of $X$ is set by some other means (e.g., the
forest catches fire due to a volcano eruption), then 
the assignment specified by $F_X$ is ``overruled''; $Y$ is
no longer committed to tracking $X$ according to $F_X$.
This emphasizes the point that variables on the left-hand side of equations
are treated differently from ones on the right-hand side.

In the philosophy community, counterfactuals are typically defined in
terms of ``closest worlds'' \cite{Lewis73,Stalnaker68}; a statement of
the form ``if $A$ were the case then $B$ would be true'' is taken to be
true if in the ``closest world(s)'' to the actual world where $A$ is true,
$B$ is also true.  This modification of equations
may be given a simple ``closest world'' interpretation:
the solution of the equations obtained by
replacing the equation for $Y$ with the equation $Y = y$,
while leaving all other equations unaltered,
gives the closest world to the actual world where $Y=y$.
(See \cite{Hal40} for further discussion of the relationship between
causal models and closest worlds.)
The asymmetry embodied in the structural equations can be understood in
terms of closest worlds.  If either the match or lightning suffice to
start the fire, then in the closest world to the actual world
where a lit match is dropped the forest burns down.  However, it is not
necessarily the case that in the world closest to one where the forest
burns down  a lit match is dropped.

It may seem somewhat circular to use
causal models, which clearly already encode causal relationships, to
define causality.  Nevertheless, as we shall see, there is no
circularity. In most examples, 
there is general agreement as to the appropriate causal model.  
The structural equations do not express actual causality; rather, they
express the effects  of interventions.  
Of course, there may be uncertainty about the effects of interventions,
just as there may be uncertainty about the true setting of the values of
the exogenous variables in a causal model.
For example, we may be uncertain about whether smoking causes cancer
(this represents uncertainty about the causal model) and uncertain about
whether a particular patient actually smoked (this is uncertainty about
the value of the exogenous variable that determines
whether the patient smokes).  This uncertainty can be described by
putting a probability on causal models and on the values of the
exogenous variables.  We can then talk about the probability that $A$ is
a cause of $B$.  (See Section~\ref{sec:blame} for further discussion of
this point.)
                                                      
In a causal model, it is possible that the value of $X$ depends on
the value of $Y$ (i.e., the equation $F_X$ is such that changes in
$Y$ can change the value of $X$) and the value of $Y$ depends on the
value of $X$.  Intuitively, this says that $X$ can potentially affect
$Y$ and that $Y$ can potentially affect $X$.  While allowed by the
framework, this type of situation does not typically happen in examples of
interest. Even when it seems to happen, we can often recast the problem
using a different (and arguably more appropriate)) choice of variables
so that we do not have such cyclic dependence (see \cite{HP01b} for an
example).  Since dealing with such cyclic dependence 
would complicate the exposition, 
I restrict attention here to what are called {\em
recursive\/} (or {\em acyclic\/}) models.  This is the special case
where there is some total ordering $\prec$ of the endogenous variables
(the ones in $\V$) 
such that if $X \prec Y$, then $X$ is independent of $Y$, 
that is, $F_X(\ldots, y, \ldots) = F_X(\ldots, y', \ldots)$ for all $y, y' \in
\R(Y)$.  Intuitively, if a theory is recursive, there is no
feedback.  If $X \prec Y$, then the value of $X$ may affect the value of
$Y$, but the value of $Y$ cannot affect the value of $X$.

It should be clear that if $M$ is an acyclic  causal model,
then given a \emph{context}, that is, a setting $\vec{u}$ for the
exogenous variables in $\U$, there is a unique solution for all the
equations.%
\footnote{A note on notation: in general, there may be several exogenous
variables in a causal model, not just one.  The ``vector'' notation
$\vec{u}$ is used here and elsewhere to denote the values of all of them.
For example, if there are three exogenous variables, say
$U_1$, $U_2$, and $U_3$, and $U_1 = 0$, $U_2 = 0$, and $U_3 = 1$, then
$\vec{u}$ is $(0,0,1)$.  This vector notation is also used to describe
the values $\vec{x}$ of a collection $\vec{X}$ of endogenous variables.}
We simply solve for the variables in the order given by
$\prec$. The value of the variables that come first in the order, that
is, the variables $X$ such that there is no variable $Y$ such that $
Y\prec X$, depend only on the exogenous variables, so their value is
immediately determined by the values of the exogenous variables.  
The values of values later in the order can be determined once we have
determined the values of all the variables earlier in the order.

\commentout{
We can describe (some salient features of) a causal model $M$ using a
{\em causal network}.%
\footnote{These causal networks are similar in spirit to the Bayesian
networks used to represent and reason about dependences in probability
distributions \cite{Pearl}.}
Figure~\ref{fig1-new} describes the causal network for
the forest-fire example.   The fact that there is an edge from $U$ to
both $L$ and $\ML$ says that the value of the exogenous variable $U$
affects the value of $L$ and $\ML$, but nothing else affects it.  The
arrows from $L$ and $\ML$ to $\FF$ say that the only the values of $\ML$
and $L$ affect the value of $\FF$.  The causal network for an acyclic
causal model is \emph{acyclic}; it has no cycles.  That is, there is no
sequence of arrows that both starts and ends at the same node. 
\begin{figure}[htb]
\input{psfig}
\centerline{\includegraphics{fig1-new}}
%
\caption{A simple causal network.}
\label{fig1-new}
\end{figure}
}

There are many nontrivial decisions to be made when choosing
the structural model to describe a given situation.
One significant decision is the set of variables used.
As we shall see, the events that can be causes and those that can be
caused are expressed in terms of these variables, as are all the
intermediate events.  The choice of variables essentially determines the
``language'' of the discussion; new events cannot be created on the
fly, so to speak.  In our running example, the fact that there is no
variable for unattended campfires means that the model does not allow us
to consider unattended campfires as a cause of the forest fire.

Once the set of variables is chosen, the next step is to decide which are
exogenous and which are endogenous.  As I said earlier, 
the exogenous variables to some extent encode the background situation
that we want to take for granted.  Other implicit background
assumptions are encoded in the structural equations themselves.
Suppose that we are trying to decide whether a lightning bolt or a
match was the
cause of the forest fire, and we want to take for granted that there is
sufficient oxygen in the air and the wood is dry.
We could model the dryness of the wood by an
exogenous variable $D$ with values $0$ (the wood is wet) and 1 (the wood
is dry).%
\footnote{Of course, in practice, we may want to allow $D$ to have more
values, indicating the degree of dryness of the wood, but that level of
complexity is unnecessary for the points that I am trying to make here.}
By making $D$ exogenous, its value is assumed to be given and out of the
control of the modeler.
We could also take the amount of oxygen bo be described by an exogenous
variable 
(for example, there could be a variable $O$ with two values---0, for
insufficient oxygen, and 1, for sufficient oxygen); alternatively, we
could choose not to model oxygen explicitly at all.  For example,
suppose that we have, as
before, a random variable $\ML$ for match dropped by arsonist,
and another variable $\WB$ for wood burning,
with values 0 (it's not) and 1 (it is).  The structural equation
$F_{\WB}$ would describe the dependence of $\WB$ on $D$
and $\ML$.  By setting $F_{\WB}(1,1) = 1$, we are saying
that the wood will burn if the lit match is dropped and the wood is dry.
Thus, 
the equation is
implicitly modeling our assumption that there is sufficient oxygen for
the wood to burn.

According to the definition of actual cause in 
Section~\ref{sec:actcaus}, only endogenous variables
can be causes or be caused.  Thus, if no variables
encode the presence of oxygen, or if it is encoded only in an exogenous
variable, then oxygen cannot be a cause of the forest burning.
If we were to explicitly model the amount of oxygen in the air (which
certainly might be relevant if we were analyzing fires on Mount
Everest), then $F_{\WB}$ would also take values of $O$ as an argument,
and the presence of sufficient oxygen might well be a cause of the wood
burning, and hence the forest burning.%
\footnote{Of course, $F_{\WB}$ might take yet other variables as
arguments.}
Interestingly, in the law, there is a distinction between what are
called \emph{conditions} and \emph{causes} \cite{Katz87}.  Under typical
circumstances, the presence
of oxygen would be considered a condition, and would thus not count as a
cause of the forest burning, while the lightning would.  While the
distinction is considered important, it does not seem to have been
carefully formalized.  One way of understanding it is in terms of
exogenous vs.~endogenous variables; conditions are exogenous, (potential)
causes are endogenous.  I discuss a complementary approach
to understanding it, in terms of theories of normality, in
Section~\ref{sec:final}.


It is not always straightforward to decide what
the ``right'' choice of variables is, or, more generally, what 
the ``right'' causal model is in a given situation; nor is it 
always obvious which of two causal models is ``better'' in some sense.
These decisions  often lie at the heart of determining
actual causality in the real world.  Disagreements about causal
relationships often boil down to disagreements about the causal model.
While the formalism presented here does not provide techniques to settle
disputes about which causal model is the right one, at least it provides
tools for carefully describing the differences between causal models, 
so that it should lead to more informed and principled decisions
about those choices (see \cite{Hal44,HH10} for more discussion of
these points). 

\section{A Formal Definition of Actual Cause}\label{sec:actcaus}

\subsection{A language for describing causes}
To make the definition of actual
causality precise, it is helpful to have a formal language for making
statements about causality.  The language is a slight extension of
\emph{propositional logic}, now often taught in highschool.  
Given a signature $\S = (\U,\V,\R)$, a \emph{primitive event} is a
formula of the form $X = x$, for  $X \in \V$ and $x \in \R(X)$.  That is,
$X$ is an endogenous variable, and $x$ is a possible value of $X$.
The primitive event $\ML=0$ says ``the lit match is not dropped''; the
primitive event $L=1$ says ``lightning occurred''.  As in propositional
logic, the symbols $\land$, $\lor$, and $\neg$ are used to denote
conjunction, disjunction, and negation, respectively.  Thus, the formula
$\ML=0 \lor L=1$ says ``either the lit match is not dropped or lightning
occurred'', $\ML=0 \land L=1$ says ``either the lit match is not dropped
and lightning occurred'', and $\neg (L=1)$ says ``lightning did not
occur'' (which is equivalent to $L=0$, given that the only possible
values of $L$ are 0 or 1).  A \emph{Boolean combination} of primitive
events is a formula 
that is obtained by combining primitive events using $\land$, $\lor$,
and $\neg$.  Thus, $\neg(\ML=0 \lor L=1) \land \WB=1$ is a Boolean
combination of the primitive events $\ML=0$, $L=1$, and $\WB=1$.

A {\em causal formula (over $\S$)\/} is one of the form
$[Y_1 \gets y_1, \ldots, Y_k \gets y_k] \phi$,
where
\begin{itemize}
\item $\phi$ is a Boolean
combination of primitive events,
\item $Y_1, \ldots, Y_k$ are distinct variables in $\V$, and
\item $y_i \in \R(Y_i)$.
\end{itemize}
Such a formula is
abbreviated
as $[\vec{Y} \gets \vec{y}]\phi$.
The special
case where $k=0$
is abbreviated as
$\phi$.
Intuitively,
$[Y_1 \gets y_1, \ldots, Y_k \gets y_k] \phi$ says that
$\phi$ would hold if
$Y_i$ were set to $y_i$, for $i = 1,\ldots,k$.

A causal formula $\psi$ is true or false in a causal model, given a
context.
I write $(M,\vec{u}) \sat \psi$  if
the causal formula $\psi$ is true in
causal model $M$ given context $\vec{u}$.
Perhaps not surprisingly, $(M,\vec{u}) \sat \psi$ is read ``$\psi$ is
true in context $\vec{u}$ in causal model model $M$''.
$(M,\vec{u}) \sat \mbox{$[\vec{Y} \gets \vec{y}](X = x)$}$ if
the variable $X$ has value $x$ in the
unique (since we are dealing with acyclic models) solution
to the equations in
$M_{\vec{Y} \gets \vec{y}}$ in context $\vec{u}$ (i.e., the
unique vector
of values for the endogenous variables that simultaneously satisfies all
equations in $\F_{\vec{Y} \gets \vec{y}}$ 
with the variables in $\U$ set to $\vec{u}$).
$(M,\vec{u}) \sat
[\vec{Y} \gets \vec{y}]\phi$ for an arbitrary Boolean combination
$\phi$ of formulas of the form $\vec{X} = \vec{x}$ is defined
similarly.  
I write $M \sat \phi$ if $(M,\vec{u}) \sat \phi$ for all contexts $\vec{u}$.

For example, if $M$ is the causal model described earlier for the forest
fire, 
then $(M,(1,1,1)) \sat [\ML \gets 0](\FF=1)$, since even if the arsonist is somehow
prevented from dropping the match, the forest burns (thanks to the
lightning);  similarly, $(M,(1,1,1)) \sat [L \gets 0](\FF=1)$.  However,
$(M,(1,1,1)) \sat [L \gets 0;\ML \gets 0](\FF=0)$: if arsonist does not drop
the lit match 
and the lightning does not strike, then forest does not burn.
Moreover, $(M,(1,1,2)) \sat \mbox{$[\ML \gets 0](\FF=1)$}$; if both
lightning and 
the match are needed for the forest to burn down, then if the arsonist
does not drop the match, the forest does not burn down.

\subsection{A preliminary definition of causality}
The HP definition of causality, like many others, is based on
counterfactuals.  The idea is that $A$ is a cause of $B$ if, if $A$
hadn't occurred (although it did), then $B$ would not have occurred.
This idea goes back to at least Hume \citeyear[Section
{VIII}]{hume:1748}, who said:
\begin{quote}
We may define a cause to  be an object followed
by another, \ldots, if the first object had not been, the second
never had existed.
\end{quote}
This is essentially the \emph{but-for} test, perhaps the most widely
used test of actual causation in tort adjudication.  The but-for test
states that an act is a cause of injury if and only if, but for the act
(i.e., had the the act not occurred), the injury would not have
occurred.

There are two well-known problems with this definition.  The first can
be seen by considering the forest-fire example again.  Suppose that an
arsonist drops a match and lightning strikes, and either one suffices to
start the fire.  Which is the cause?
According to a naive interpretation of the counterfactual definition,
neither is.  If the match hadn't dropped, then the lightning would
still have struck, so there would have been a forest fire anyway.
Similarly, if the lightning had not occurred, there still would have
been a forest fire.  Our definition will declare both lightning and the
arsonist cases of the fire.  (In general, there may be more than one
cause of an outcome.)

A more subtle problem is what philosophers have called
\emph{preemption}, where there are two potential causes
of an event, one of which preempts the other.  Preemption 
is illustrated by the following story, taken from \cite{Hall98}:

\begin{quote}
Suzy and Billy both pick up rocks
and throw them at  a bottle.
Suzy's rock gets there first, shattering the
bottle.  Since both throws are perfectly accurate, Billy's would have
shattered the bottle had it not been preempted by Suzy's throw.
\end{quote}
Common sense suggests that Suzy's throw is the cause of the shattering,
but Billy's is not.  However, Suzy's throw is not a cause according to
the naive counterfactual definition; if Suzy hadn't thrown, then Billy's
throw would have 
shattered the bottle.

The HP definition deals with the first  problem by defining causality as 
counterfactual dependency \emph{under certain contingencies}.
In the forest-fire example, the forest fire does counterfactually depend
on the lightning under the contingency that the arsonist does not drop
the match; similarly, the forest fire depends counterfactually on the
arsonist's match under the contingency that the lightning does not
strike.  Clearly we need to be a little careful here to limit the
contingencies that can be considered.  We do not want to make Billy's
throw the cause of the bottle shattering by considering the contingency
that Suzy does not throw.   The reason that we consider Suzy's throw to
be the cause and Billy's throw not to be the cause is that Suzy's rock
hit the bottle, while Billy's did not.  Somehow the definition must
capture this obvious intuition.

With this background, HP's preliminary definition
of causality can be given; it is 
quite close to the final definition, given in Section~\ref{sec:final}.
The types of events that the HP definition allows as actual causes are
ones of the form $X_1 = x_1 \land \ldots \land X_k = x_k$---that is,
conjunctions of primitive events; this is often abbreviated as $\vec{X}
= \vec{x}$. The events that can be caused are arbitrary Boolean
combinations of primitive events.
The definition does not allow statements of the form  ``$A$ or $A'$ is a
cause of $B$,'' although this could be treated as being equivalent to
``either $A$ is a cause of $B$ or $A'$ is a cause of $B$''.    
On the other hand, statements such as
``$A$ is a cause of $B$ or $B'$'' are allowed;  
this is not
equivalent to ``either $A$ is a cause of $B$ or $A$ is a cause of $B'$''.

\dfn\label{actcaus}
(Actual cause; preliminary version)
$\vec{X} = \vec{x}$ is an {\em actual cause of $\phi$ in
$(M, \vec{u})$ \/} if the following
three conditions hold:
\begin{description}
\item[{\rm AC1.}]\label{ac1} $(M,\vec{u}) \sat (\vec{X} = \vec{x})$ and 
$(M,\vec{u}) \sat \phi$.
\item[{\rm AC2.}]\label{ac2}
There is a partition of $\V$ (the set of endogenous variables) into two
subsets $\vec{Z}$ and $\vec{W}$  
with $\vec{X} \subseteq \vec{Z}$ and a
setting $\vec{x}'$ and $\vec{w}$ of the variables in $\vec{X}$ and
$\vec{W}$, respectively, such that
if $(M,\vec{u}) \sat Z = z^*$ for 
all $Z \in \vec{Z}$, then
both of the following conditions hold:
\begin{description}
\item[{\rm (a)}]
$(M,\vec{u}) \sat [\vec{X} \gets \vec{x}',
\vec{W} \gets \vec{w}]\neg \phi$.
\item[{\rm (b)}]
$(M,\vec{u}) \sat [\vec{X} \gets
\vec{x}, \vec{W}' \gets \vec{w}, \vec{Z}' \gets \vec{z}^*]\phi$ for 
all subsets $\vec{W}'$ of $\vec{W}$ and all subsets $\vec{Z'}$ of
$\vec{Z}$, where I abuse notation and write $\vec{W}' \gets \vec{w}$ to
denote the assignment where the variables in $\vec{W}'$ get the same
values as they would in the assignment $\vec{W} \gets \vec{w}$
(thus, components in the vector $\vec{w}$ that do not match any
variable in $\vec{W}'$ are ignored).
\end{description}
\item[{\rm AC3.}] \label{ac3}
$\vec{X}$ is minimal; no subset of $\vec{X}$ satisfies
conditions AC1 and AC2.\ \ \ \bbox
\label{def3.1}  
\end{description}
\end{definition}

AC1 is just says that $\vec{X}=\vec{x}$ cannot
be considered a cause of $\phi$ unless both $\vec{X} = \vec{x}$ and
$\phi$ actually happen.  AC3 is a minimality condition, which ensures
that only those elements of 
the conjunction $\vec{X}=\vec{x}$ that are essential for
changing $\phi$ in AC2(a) are
considered part of a cause; inessential elements
are pruned. 
Without AC3, if dropping a lit match qualified as a
cause of the forest fire, then dropping a match and
sneezing would also pass the tests of AC1 and AC2.
AC3 serves here to strip ``sneezing''
and other irrelevant, over-specific details
from the cause.  

Clearly, all the ``action'' in the definition occurs in AC2.
We can think of the variables in $\vec{Z}$ as making up the ``causal
path'' from $\vec{X}$ to $\phi$.  Intuitively, changing the value of
some variable in $\vec{X}$ results in changing the value(s) of some
variable(s) in $\vec{Z}$, which results in the values of some
other variable(s) in $\vec{Z}$ being changed, which finally results in
the value of $\phi$ changing.  The remaining endogenous variables, the
ones in $\vec{W}$, are off to the side, so to speak, but may still have
an indirect effect on what happens.  AC2(a) is essentially the standard
counterfactual definition of causality, but with a twist.  If we 
want to show that $\vec{X} = \vec{x}$ is a cause of $\phi$, we must show
(in part) that if $\vec{X}$ had a different value, then so too would
$\phi$.  However, this effect of $\vec{X}$ on the value of
$\phi$ may not hold in the actual context; 
the value of $\vec{W}$ may have to be different to allow the effect to
manifest itself.  For example, in the context where both the lightning
strikes and the arsonist drops a match, stopping the arsonist from
dropping the match will not prevent the forest fire.  The
counterfactual effect of the arsonist on the forest fire manifests
itself only in a situation where the lightning does not strike (i.e., where
$L$ is set to 0).  

AC2(b) is perhaps the most complicated condition.  It limits the
``permissiveness'' of AC2(a) with regard to the 
contingencies (i.e., the 
values of the variables in
$\vec{W}$) that can be considered.  Essentially, it ensures that
$\vec{X}$ alone suffices to bring about the change from $\phi$ to $\neg
\phi$; setting $\vec{W}$ to $\vec{w}$ merely eliminates
possibly spurious side effects that may mask the effect of changing the
value of $\vec{X}$.  Moreover, although the values of variables on the
casual path (i.e.,  the variables $\vec{Z}$) may be perturbed by
the change to $\vec{W}$, this perturbation has no impact on the value of
$\phi$.  Note that if  $(M,\vec{u}) \sat \vec{Z} = \vec{z}^*$, then $\vec{z}^*$ is
the value of the 
variables in $\vec{Z}$ in the context $\vec{u}$.  We capture the fact
that the perturbation has no impact on the value of $\phi$ by saying
that if some variables $\vec{Z}'$ on the causal path were set to their
original values in the context $\vec{u}$, $\phi$ would still be true, as
long as $\vec{X} = \vec{x}$.

This condition is perhaps best understood by considering the
Suzy-Billy example.  It is an attempt to capture the intuition that the
reason that Suzy is the cause of the bottle shattering, and not Billy,
is that Suzy's rock actually hit the bottle and Billy's didn't.  
If there is a variable in the model that represents whether Billy's rock
hits the bottle, then for Billy's throw to be a cause of the bottle
shattering, the bottle would have to shatter (that will be the $\phi$ in
AC2(b)) if Billy throws ($X=x$) and Suzy does not ($W = w$) even if, as
is actually the case in the real world, Billy's rock does not hit the
bottle ($Z = z^*$).  This should become clearer when we analyze this
example more formally, in Example~\ref{xam2}.   Before considering this
examples, I start with a simpler example, just to give the intuitions.  

\commentout{
The version of AC2(b) in \cite{HP01b} is slightly different
from that in the preliminary version of the paper \cite{HPearl01a}.
For the original definition, it was shown that the minimality condition
AC3 guarantees that causes are always single conjuncts 
 \cite{EL01,Hopkins01}.  It was claimed in \cite{HP01b} that the result
was still true for the modified definition, but Example~\ref{xam3b}
shows this is not the case.
Nevertheless, it is shown that it is true for most cases of
interest (cf.~Proposition~\ref{pro:singlecause}).  Before considering these
examples, I start with a simpler example, just to give the intuitions.  
}

\xam\label{ex:ff}
For the forest-fire example, first consider the context where $U =
(1,1,1)$, so that the lightning strikes, the arsonist drops the match,
and either one suffices to start the fire.  
both the dropped match and the lightning are causes of the forest fire
in this context.  Here is the argument for lightning (the
argument for the dropped match is symmetric).

Let $M$ be the causal model for the forest fire described earlier,
where the endogenous variables are $L$, $\ML$, and $\FF$, and $U$ is the
only exogenous variable.  In Definition~\ref{actcaus}, $\vec{X} =
\vec{x}$ is $L=1$ and $\phi$ is $\FF=1$.  
Clearly $(M,(1,1,1)) \sat \FF=1$ and 
$(M,(1,1,1)) \sat L=1$; in the context (1,1,1), the lightning strikes
and the forest burns down.  Thus, AC1 is satisfied.  AC3 is trivially
satisfied, since $\vec{X}$ consists of only one element, $L$, so must be
minimal.  For AC2, let $\vec{Z} = \{L, \FF\}$, $\vec{W} =
\{\ML\}$, $x' = 0$, and $w = 0$.  Clearly,
$(M,(1,1,1)) \sat \mbox{$[L \gets 0, \ML \gets 0](\FF \ne 1)$}$; if
the lightning 
does not strike and the match is not dropped, the forest does not burn
down, so AC2(a) is satisfied.  To see the effect of the lightning, we
must consider the contingency where the match is not dropped; the
definition allows us to do that by setting $\ML$ to 0.  (Note that 
setting $L$ and $\ML$ to 0 overrides the effects of $U$; this is
critical.)  Moreover,  
$(M,(1,1,1)) \sat [L \gets 1, \ML \gets 0](\FF = 1)$;
if the lightning
strikes, then the forest burns down even if the lit match is not
dropped, so AC2(b) is satisfied.  (Note that since $\vec{Z} = \{L, \FF\}$, 
the only subsets of $\vec{Z} - \vec{X}$ are the empty set and the
singleton set consisting of just $\FF$.)

The lightning and the dropped match are also causes of the forest fire
in the context $(1,1,2)$, where both the lightning and 
match are needed to start the fire.  
Again, I just present the argument for the lightning here.
As before, both AC1 and AC3 are trivially satisfied.  For AC2, again
take $\vec{Z} = \{L, \FF\}$, $\vec{W} =
\{\ML\}$, and $x' = 0$, but now let  $w = 1$.
We have that 
$$\begin{array}{c}
(M,(1,1,2)) \sat [L \gets 0, \ML \gets 1](\FF \ne 1) \mbox{ and }\\
(M,(1,1,2)) \sat [L \gets 1, \ML \gets 1](\FF = 1),\end{array}$$
so AC2(a) and AC2(b) are satisfied.  

As this example shows, causes are not
unique; there may be more than one cause of a given outcome.  Moreover,
both the lightning and the dropped match are causes both 
in the case where either one suffices to start the fire and in the
case where both are needed.  As we shall see, the notion of
\emph{responsibility} distinguishes these two situations.  Finally, it
is worth noting that the lightning is not the cause in either the context
$(1,0,2)$ or the context $(1,1,0)$.  In the first case, AC1 is violated.
If both the lightning and the match are needed to cause the fire, then
there is no fire if the match is not dropped.   In the second case,
there is a fire but, intuitively, it arises spontaneously; neither the
lightning nor the dropped match are needed.  Here AC2(a) is violated;
there is no setting of $L$ and $\ML$ that will result in no forest fire.
\exam

\xam\label{xam2}
Now let us consider the Suzy-Billy example.%
\footnote{The discussion of this and the following example is taken
almost verbatim from HP.}
We get the desired result---that Suzy's throw is a cause, but Billy's is
not---but only if we 
model the story appropriately.  Consider first a coarse causal
model, with three endogenous variables:
\begin{itemize}
\item $\ST$ for ``Suzy throws'', with values 0 (Suzy does not throw) and
1 (she does);
\item $\BT$ for ``Billy throws'', with values 0 (he doesn't) and
1 (he does);
\item $\BS$ for ``bottle shatters', with values 0 (it doesn't shatter)
and 1 (it does).
\end{itemize}
(I omit the exogenous variable here; it determines whether Billy and
Suzy throw.)  Take the formula for $\BS$ to be such that the bottle
shatters if either Billy or Suzy throw; that is $\BS = \BT \vee \ST$. (I
am implicitly assuming that Suzy and Billy never miss if they throw.)
$\BT$ and $\ST$ play symmetric roles in this model;
there is nothing to distinguish them.
Not surprisingly, both
Billy's throw and Suzy's throw are classified as causes of the
bottle shattering
in this model.  The argument is essentially identical to the forest-fire
example in the case that $U = (1,1,1)$, where either the lightning or
the dropped match is enough to start the fire.

The trouble with this model is that it cannot distinguish
the case where both rocks
hit the bottle simultaneously (in which case it would be reasonable
to say that both $\ST=1$ and $\BT=1$ are 
causes of $\BS=1$) from the case where Suzy's rock
hits first.  The model has to be refined to express this distinction.
One way is to invoke a dynamic model \cite[p.~326]{pearl:2k}.  This
model is discussed by HP.
A perhaps simpler way to gain expressiveness is to allow $\BS$ to be
three valued, with values 0 (the bottle doesn't shatter), 1
(it shatters as a result of being hit by Suzy's rock), and 2 (it
shatters as a result of being hit by Billy's rock).
I leave it to the reader to check that $\ST = 1$ is a
cause of $\BS = 1$, but $\BT = 1$ is not (if Suzy doesn't thrown but
Billy does, then we would have $\BS = 2$).  Thus, to some extent, this
solves our problem.  But it
borders on cheating; the answer is almost
programmed into the model by invoking the relation ``as a result of'',
which requires the identification of the actual cause.

A more useful choice is to add two new random variables to the model:
\begin{itemize}
\item $\BH$ for ``Billy's rock hits the (intact) bottle'', with values 0
(it doesn't) and 1 (it does); and
\item $\SH$ for ``Suzy's rock hits the bottle'', again with values 0 and
1.
\end{itemize}
Now it is the case that, in the context where both Billy and Suzy throw,
$\ST=1$ is a cause 
of $\BS=1$, but $\BT = 1$ is not.
To see that $\ST=1$ is a cause, note that, as usual, it is immediate
that AC1 and AC3 hold.  For AC2, choose $\vec{Z} = \{\ST,\SH,\BH\}$,
$\vec{W}=\{\BT\}$,  and $w=0$. 
When $\BT$ is set to 0, $\BS$ tracks $\ST$: if Suzy
throws, the bottle shatters  and if she doesn't throw, the bottle does
not shatter.  To see that $\BT=1$ is \emph{not} a cause
of $\BS=1$, we must check that there is no
partition $\vec{Z} \cup \vec{W}$ of the endogenous variables that
satisfies AC2.
Attempting the symmetric choice with $\vec{Z} = \{\BT,\BH,\SH\}$,
$\vec{W}=\{\ST\}$, and $w=0$ 
violates AC2(b). To see this, take $\vec{Z}' = \{\BH\}$.  In the
context where Suzy 
and Billy both throw, $\BH=0$.  If $\BH$ is set to 0, the bottle does
not shatter if Billy throws and Suzy does not.  
It is precisely because, in this context, Suzy's throw hits the bottle
and Billy's does not that we declare Suzy's throw to be the cause of the
bottle shattering.  AC2(b) captures that intuition by allowing us to
consider the contingency where $\BH=0$, despite the fact that Billy
throws.   I leave it to the reader to check that no other partition of
the endogenous variables satisfies AC2 either.  


This example emphasizes an important moral.
If we want to argue in a case of preemption
that $X=x$ is the cause of $\phi$ rather than $Y=y$,
then there must be a random variable ($\BH$ in this case) that takes on
different values depending on whether $X=x$ or $Y=y$ is the actual
cause.  If the model does not contain such a variable, then it will not
be possible to determine which one is in fact the cause.  This is
certainly consistent with intuition and the way we present evidence.  If
we want to  argue (say, in a court of law) that it was $A$'s shot that
killed $C$ rather than $B$'s, then we present evidence such as the
bullet entering $C$ from the left side (rather than the right side, which
is how it would have entered had $B$'s shot been the lethal one).
The side from which the shot entered is the relevant random variable in
this case.  Note that the random variable may involve temporal evidence
(if $Y$'s shot had been the lethal one, the death would have occurred
a few seconds later), but it certainly does not have to.
\exam

\xam\label{xam4}  Can {\em not} performing an action be (part of) a cause?
Consider the following story, also taken from
(an early version of) \cite{Hall98}:

\begin{quote}
Billy, having stayed out in the cold too long throwing rocks, contracts a
serious but nonfatal disease.  He is hospitalized
and treated on Monday, so is fine Tuesday morning.
\end{quote}

But now suppose the doctor does not treat Billy on Monday.
Is the doctor's omission to treat Billy a cause of Billy's
being sick on Tuesday?  It seems that it should be, and indeed it is
according to our analysis.  Suppose that $\vec{u}$ is the context where,
among other things, Billy is sick on Monday and the situation is such
that the doctor forgets to administer the medication Monday.
It seems reasonable that
the model should have two random variables:
\begin{itemize}
\item $\MT$ for ``Monday treatment'', with values 0 (the doctor does
not treat Billy on Monday) and 1 (he does); and
\item $\BMC$ for ``Billy's medical condition'', with values 0 (recovered)
and 1 (still sick).
\end{itemize}
Sure enough, in the obvious causal model, $\MT=0$ is a cause
of $\BMC=1$.

This may seem somewhat disconcerting at first.  Suppose there are 100
doctors in the hospital.  Although only one of them was assigned to
Billy (and he forgot to give medication), in principle, any of the other
99 doctors could have given Billy his medication.  Is the fact that they
didn't give him the medication also part of the cause 
of him still being sick on Tuesday?

In the causal model above, the other doctors' failure to give Billy his
medication is not a cause, since the model has no random variables to
model the other doctors' actions, just as there was  no random variable
in the causal model of Example~\ref{ex:ff} to model the presence of oxygen. 
Their lack of action is part of the context.  We factor it out because
(quite reasonably) we want to focus on the actions of Billy's doctor.
If we had included endogenous random variables corresponding to the
other doctors, then they too would be causes of Billy's
being sick on Tuesday.  The more refined definition of causality
given in the next section provides a way of avoiding this
problem even if the model includes endogenous variables for the other
doctors.   

With this background, I
continue with Hall's modification of the original story.

\begin{quote}
Suppose that Monday's doctor is reliable, and administers the medicine
first thing in the morning, so that Billy is fully recovered by Tuesday
afternoon.  Tuesday's doctor is also reliable, and would have treated
Billy if Monday's doctor had failed to.  \ldots And let us add a twist:
one dose of medication is harmless, but two doses are lethal.
\end{quote}
Is the fact that Tuesday's doctor did {\em not\/}
treat Billy the cause of him being alive (and recovered) on
Wednesday morning?

The causal model for this story is straightforward.
There are three random variables: 
\begin{itemize}
\item $\MT$ for Monday's treatment (1 if
Billy was treated Monday; 0 otherwise); 
\item $\TT$ for Tuesday's treatment (1
if Billy was treated Tuesday; 0 otherwise); and 
\item $\BMC$ for Billy's
medical condition
(0 if Billy is fine both Tuesday morning and Wednesday morning;
1  if Billy is
sick Tuesday morning, fine Wednesday morning; 2 if Billy
is sick both Tuesday and Wednesday morning; 3  if Billy is fine
Tuesday morning and dead Wednesday morning).
\end{itemize}
We can then describe Billy's condition as a function of the four
possible combinations of treatment/nontreatment on Monday and Tuesday.
I omit the obvious structural equations corresponding to this
discussion.  

In this causal model, it is true that $\MT=1$ is a cause
of $\BMC=0$, as we would expect---because Billy is treated Monday, he is not
treated on Tuesday morning, and thus recovers Wednesday morning.
$\MT=1$ is also a cause
of $\TT=0$, as we would
expect, and $\TT=0$ is a cause
of Billy's being alive ($\BMC=0
\lor \BMC=1 \lor \BMC=2$).  However, $\MT=1$ is {\em not\/} a cause
of Billy's being alive.  It fails condition AC2(a): setting
$\MT=0$ still leads to Billy's being alive (with $W=\emptyset$).
Note that it would not help to take $\vec{W}= \{\TT\}$.  For if $\TT=0$,
then Billy is alive no matter what $\MT$ is, while if $\TT=1$, then Billy is
dead when $\MT$ has its original value, so AC2(b) is violated (with
$\vec{Z}' = \emptyset$).

This shows that
causality is not transitive, according to our definitions.
Although $\MT=1$ is a cause of $\TT=0$ and $\TT=0$ is a
cause of $\BMC=0 \lor \BMC=1 \lor \BMC=2$, $\MT=1$ is not a cause
of $\BMC=0 \lor \BMC=1 \lor \BMC=2$.
Nor is causality closed
under {\em right weakening}:  $\MT=1$ is a cause of $\BMC=0$,
which logically implies $\BMC=0 \lor \BMC=1 \lor \BMC=2$, which is not
caused by $\MT=1$.  

This distinguishes the HP definition from that of Lewis
\citeyear{Lewis00}, which builds in
transitivity and implicitly assumes right weakening.  
\exam

\commentout{
The version of AC2(b) used here is taken from \cite{HP01b}, and differs
the version given in the conference version of that paper \cite{HPearl01a}.
Here, AC2(b) is required to hold for all subsets $\vec{W}'$ of
$\vec{W}$; in the original definition, it was required to hold only for
$\vec{W}$.  The following example, due to Hopkins and Pearl
\citeyear{HopkinsP02}, illustrates why the change was made.

\xam\label{xam3} Suppose that a prisoner dies 
either if $A$ loads $B$'s gun and $B$ shoots, or if $C$ loads and shoots
his gun.  Taking $D$ to represent the prisoner's death and making the
obvious assumptions about the meaning of the variables, we have that
$D=1$ iff $(A=1 \land B=1) \lor (C=1)$.  Suppose that in the actual
context $u$, $A$ loads $B$'s gun, $B$ does not shoot, but $C$ does load
and shoot his gun, so that the prisoner dies.  Clearly $C=1$ is a cause
of $D=1$.  We would not want to say that $A=1$ is a cause of $D=1$ in
context $u$; given that $B$ did not shoot (i.e., given that $B=0$),
$A$'s loading the gun should not count as a cause.  The obvious way to
attempt to show that $A=1$ is a cause is to take $\vec{W} = \{B,C\}$ and
consider the contingency where $C=0$ and $B=1$.
It is easy to check that AC2(a) holds for this contingency; moreover,
$(M,u) \sat [A \gets 1, B \gets 1, C \gets 0](D=1)$.  However,
$(M,u) \sat [A \gets 1, C \gets 0](D=0)$.  Thus, AC2(b) is not satisfied
for the subset $\{C\}$ of $W$, so $A=1$ is not a cause of $D=1$.  However,
had we required AC2(b) to hold only for $\vec{W}$ rather than all
subsets $\vec{W}'$ of $\vec{W}$, then $A=1$ would have been a cause.
\exam

\commentout{
It turns out that taking $\vec{W}$ to be $\V - \vec{X}$ (as in
\cite{HP01b}) or a subset of $\V- \vec{X}$ (as was done here) makes no
difference in the original definition.  However, as the following
example shows, it does make a difference once AC2(b) is required to hold
for all subsets of $\vec{W}$. Moreover, the current version of AC2(b)
seems to give more reasonable answers.

\xam\label{xam3a}.
$A$ and $B$ both vote for a candidate.  One vote is
all it takes to win.  In the actual context $u$, both vote for him.  In this
case, an analysis just like that in the forest-fire example shows that
$A=1$ and $B=1$ are both causes of $\WIN=1$.  However, now there is a
twist.  Because the voting machines are considered unreliable, each
voters votes are recorded both by  a paper ballot and by an optical
scan.  If the paper ballot and the voter scan agree, the vote counts.
Otherwise, the election is declared invalid.  Suppose that $A_1$ and
$A_2$ record $A$'s votes, and $B_1$ and $B_2$ record $B$'s votes.
Thus, we have the equations $A_1 \gets A$, $A_2 \gets A$, $B_1 \gets B$,
and $B_2 \gets B$. Finally, we have $\WIN = 1$ iff ($A_1 = A_2 = 1$ and
$B_1 = B_2$) or $(B_1 = B_2 = 1 and A_1 = A_2)$.  We would like to say
that $A = 1$ and $B = 1$ are each causes of $\WIN=1$.  That is the case
with the definition in this paper.  To see that $A = 1$ is a cause, take
$\vec{Z} = \{A, A_1, A_2,\WIN\}$ and 
$\vec{W} = \{B\}$.  It is easy to see that if $A=0$ and $B=0$, then
$\WIN=0$, while if $A=1$, then $W=1$ whether or not we set $B=0$.  A
similar argument shows that $B=1$ is a cause of $\WIN=1$.  

However, this argument does not work if the definition in \cite{HP01b}.
For suppose that we do the obvious thing, which is to take $\vec{Z} =
\{A, A_1, A_2,\WIN\}$ and $\vec{W} = \{B,B_1,B_2\}$, and consider the
obvious contingency where $B=B_1=B_2 = 0$.  Indeed, it is the case that,
under this contingency, if $A=1$, then $\WIN=1$, and if $A=0$, then
$\WIN=1$.  However, setting $A=1$ does not force $\WIN=1$ for all
subsets $\vec{W}'$ of $\vec{W}$.  For example, if  $\vec{W}' = \{B_2\}$,
then $B = B_1 = 1$, since this is what happens in the
actual context, so $\WIN=0$, even if $A=1$.  I leave it to the reader to
check that no other partition shows that $A=1$ is a cause of $\WIN=1$.
Moreover, with the version of AC2(b) in \cite{HP01b}, $A = 1 \land B =
1$ is a cause of $\WIN=1$ (taking $\vec{Z} = \V$ and $\vec{W} =
\emptyset$), showing that causes do not have to be singletons.
\exam
}

While the change in AC2(b) has the advantage of being able to deal with
Example~\ref{xam3}, it has a nontrivial side effect.   
For the original definition, it was shown that the minimality condition 
AC3 guarantees that causes are always single conjuncts 
\cite{EL01,Hopkins01}.  It was claimed in \cite{HP01b} that the result
was still true for the modified definition, but, as I now show, this is
not the case. 

\xam\label{xam3b} $A$ and $B$ both vote for a candidate.  $A$'s vote is
recorded in an optical scanner ($C$).  If $B$ votes for the candidate,
then she wins; if $A$ votes for the candidate and her vote agrees with
optical scanner vote, then the candidate wins.  If $B$ does not vote for 
the candidate, then $A$ will not either.  Unfortunately, $B$
also has access to the scanner, so she will set it to read 0 if she does
not vote for the candidate.  In the actual context $\vec{u}$, both $A$
and $B$ vote for the candidate.   The following structural equations
characterize $C$ and $\WIN$: $C=\min(A,B)$, and $\WIN = 1$ iff $B=1$ or
$A=C=1$.   I now show that $A=1 \land C=1$ is a cause of $\WIN=1$,
but neither $A=1$ nor $C=1$ is a cause.  To see that $A=1 \land C=1$ is
a cause, first observe that AC1 clearly holds.  For AC2, let $\vec{W} =
\{B\}$ (so $\vec{Z} = \{A,C,\WIN\}$) and take $w = 0$ (so we are considering
the contingency where $B=0$).  Clearly, $(M,\vec{u}) \sat [A \gets 0, C
\gets 0, B \gets 0](\WIN=0)$ and $(M,\vec{u}) \sat [A \gets 1, C \gets
1, B \gets 0](\WIN=1)$, so AC2 holds.  To show that AC3 holds, I must
show that neither $A=1$ nor $C=1$ is a cause of $\WIN=1$.  To show that $A=1$
is not a cause, note that if $A=1$ is a cause with $\vec{W}$,
$\vec{w}$, and $\vec{x}'$ as witnesses, then  $\vec{W}$ must contain $B$
and $\vec{w}$ must be such that $B=0$.   But since $(M,u) \sat [A \gets
1, B \gets 
0](\WIN = 0)$, AC2(b) is violated no matter whether $C$ is in $\vec{Z}$
or in $\vec{W}$.  Similarly, if $C=1$ is a cause with $\vec{W}$,
$\vec{w}$, and $\vec{x}'$ as witnesses, then  $\vec{W}$ must again
contain $B$ and $\vec{w}$ must again be such that $B=0$.   But since
$(M,u) \sat [C \gets 1, B \gets 
0](\WIN = 0)$, AC2(b) is violated no matter whether $A$ is in $\vec{Z}$
or in $\vec{W}$. 
\exam

Although Example~\ref{xam3b} shows that causes are not always single
conjuncts, they often are.  Indeed, it is not hard to show that in all
the standard examples considered in the philosophy and legal literature
(in particular, in all the examples considered in HP), they are.  
The following result give some intuition as to why.  
Notice that in Example~\ref{xam3b}, $B$ affects both $A$ and $C$.  This
turns out to be the key to getting causes that are not single conjuncts.

\commentout{
\dfn 
If $M = (\U,\V,\F)$ is a causal model and 
$X \union \vec{Y} \subseteq \V$, then 
$X$ \emph{is not affected by} $\vec{Y}$ in $(M,\vec{u}$ if
$(M,\vec{u}) \sat X = x$ implies $(M,\vec{U}) \sat [\vec{Y} \gets
\vec{y}](X=x)$ for 
all possible settings $\vec{y}$ of the variables in $\vec{Y}$.  That is,
setting $\vec{Y}$ to an arbitrary value does not 
cause the value of $X$ to change in context $\vec{u}$ in $M$.
\edfn
}

Say that $\vec{X}
= \vec{x}$ is a \emph{weak cause of $\phi$ under the contingency 
$\vec{W} = \vec{w}$ in $(M,\vec{u})$} if AC1 and AC2 hold under
the contingency $\vec{W} = \vec{w}$, but AC3 does not necessarily hold.

\pro\label{pro:singlecause} If $\vec{X} = \vec{x}$ is a weak cause of
$\phi$ in $(M,\vec{u})$ with $\vec{W}$, $\vec{w}$, and $\vec{x}'$ as
witnesses, $|\vec{X}| > 1$, and each variable  $X_i$ in $\vec{X}$
is independent of all the variables in $\V - \vec{X}$, then $\vec{X} =
\vec{x}$ is not a cause of $\phi$ in $(M,\vec{u})$. 
\epro

\prf  Suppose that the hypotheses of the proposition hold.
First note that since $\vec{X} = \vec{x}$ is a weak
cause of $\phi$ in $(M,\vec{u})$, by AC1, we must have 
$(M,\vec{u}) \sat \vec{X} = \vec{x}$.  Since each variable in
$\vec{X}$ is independent of all the variables in $\V - \vec{X}$, 
for all $\vec{Y} \subseteq \V - \vec{X}$ and all settings $\vec{y}$ of
the variables in $\vec{Y}$, we must have
$(M,\vec{u}) \sat [\vec{Y} \gets \vec{y}](\vec{X} = \vec{x})$.
It follows that, for all formulas $\psi$, all subsets $\vec{X}'$ of
$\vec{X}$, all subsets $\vec{Y}$ of $\V - \vec{X}$, and all settings
$\vec{y}$ of $\vec{Y}$, we have
\begin{equation}\label{eq1}
(M,\vec{u}) \sat [\vec{Y} \gets \vec{y}]\psi \mbox{ iff }
(M,\vec{u}) \sat [\vec{X}' \gets \vec{x}, \vec{Y} \gets \vec{y}]\psi.
\end{equation}

Next, observe that since the causal model is acyclic, there must be some
variable in $\vec{X}$ that is independent of every other variable in
$\vec{X}$.  Without loss of generality, suppose that it is $X_1$.  
Thus, $X_1$ is independent of every variable in $\V - \{X_1\}$. 
Let $\vec{X}^- = \<X_2, \ldots, X_k\>$ and $\vec{x}^- = \<x_2,
\ldots, x_k\>$.  I show that either $X_1 = x_1$ or $\vec{X}^- = \vec{x}^-$
is a weak cause of $\phi$, showing that $\vec{X} =
\vec{x}$ is not a cause of $\phi$, since it does not satisfy AC3.  

First suppose that $x_1 = x_1'$.  I show that then
$\vec{X}^- = \vec{x}^-$ is a weak cause of $\phi$, with $\vec{W}$,
$\vec{w}$, and $\vec{x}'$ as witnesses.  To see this, note that since
$\vec{X} =
\vec{x}$ is a weak cause of $\phi$, with $\vec{W}$, $\vec{w}$, and
$\vec{x}'$ as witnesses, by AC2(a), we have that
$(M,\vec{u}) \sat [\vec{X} \gets \vec{x}', \vec{W}
\gets \vec{w}]\neg \phi$. 
By the same arguments as used to derive (\ref{eq1}), we have 
that $(M,\vec{u}) \sat [\vec{X}^- \gets \vec{x}^-, \vec{W} \gets \vec{w}]
\neg \phi$.  Thus, AC2(a) holds for $\vec{X}^- = \vec{x}^-$.  By AC2(b), 
$(M,\vec{u}) \sat [\vec{X} \gets \vec{x}, \vec{W}' \gets \vec{w},
\vec{Z}' \gets \vec{z}^*]\phi$ for 
all subsets $\vec{W}'$ of $\vec{W}$ and all subsets $\vec{Z}'$ of
$\vec{Z}$.  By (\ref{eq1}), we have that 
$(M,\vec{u}) \sat [\vec{X}^- \gets \vec{x}^-, \vec{W}' \gets \vec{w},
\vec{Z}' \gets \vec{z}^*]\phi$.  Thus, AC2(b) holds for  $\vec{X}^- =
\vec{x}^-$, and $\vec{X}^- = \vec{x}^-$ is indeed a weak cause of $\phi$.

Now suppose that $x_1 \ne x_1'$.   If $\vec{X}^- = \vec{x}^-$ is a weak
cause of $\phi$ with witnesses $\vec{W} \union \{X_1\}$, $\vec{w}\cdot
\<x_1'\>$, and $\vec{x}^-$, then 
we are done.  It is immediate that AC1 holds for $\vec{X}^- =
\vec{x}^-$, and that AC2(a) hold with these witnesses.  Thus, AC2(b)
must fail.  It follows that
there must exist some subset $\vec{W}'$ of
$\vec{W}$ and subset $\vec{Z}'$ of $\vec{Z}$ such that either 
(a) $(M,\vec{u}) \sat [\vec{X}^- \gets
\vec{x}^-, X_1 \gets x_1', \vec{W}' \gets \vec{w},  \vec{Z}' \gets
\vec{z}^*]\neg \phi$ or 
(b) $(M,\vec{u}) \sat [\vec{X}^- \gets
\vec{x}^-, \vec{W}' \gets \vec{w},  \vec{Z}' \gets
\vec{z}^*]\neg \phi$.  Option (b) cannot hold, because, by 
(\ref{eq1}), it holds iff $(M,\vec{u}) \sat [\vec{X} \gets
\vec{x}, \vec{W}' \gets \vec{w},  \vec{Z}' \gets
\vec{z}^*]\neg \phi$, which contradicts the assumption that $\vec{X}
\gets \vec{x}$ is a weak cause of $\phi$ with $\vec{W}$, $\vec{w}$, and
$\vec{x}'$ as witnesses.  Thus we must have (a).
But now it follows that $X_1 = x_1$ is a cause of $\phi$, with
with $\vec{W} \union \vec{X}^-$, $\vec{w} \cdot \vec{x}^-$, and $x_1'$
as witnesses: AC1 and AC3 are immediate, AC2(a) follows from the fact
that $\vec{X} = \vec{x}$ is a weak cause with $\vec{W}$, $\vec{w}$ and
$\vec{x}'$ as witnesses, and AC2(b) follows from (a) above.
\eprf
 
In the examples in \cite{HP01b} (and elsewhere in the literature), the
each variable in a potential cause $\vec{X}$ is typically independent of
the variables  $\vec{W}$ that are set in the relevant contingency, so
causes are in fact single conjuncts.
}

\section{Dealing with normality and typicality}\label{sec:final}

While the definition of causality given in Definition~\ref{actcaus}
works well in many cases, it does not always deliver answers that
agree with (most people's) intuition.  Consider the following example,
taken from Hitchcock \citeyear{Hitchcock07}, based on an example due
to Hiddleston \citeyear{Hiddleston05}.
\commentout{
\xam\label{xam:Pandu}
Pandu lives in a village where the weather is always glorious,
although there are stories that, in the time of his
great-great-grandfather, there was a tremendous wind that flattened his
house, and it had to be rebuilt.  Pandu had an excellent fishing season
last year. He spent some of the money he made reinforcing his house, so
it will never be flattened again; this impressed his neighbors.  The
next year the weather is marvelous, as usual, and his house does not
get flattened.  Is the reinforcement a cause of his house not being
flattened?

I think that just about everyone would agree that, if there were winds, then
the reinforcement was a cause of the house not being flattened.  But in
the absence of wind, most people would say that it is not a cause.
Nevertheless, in the obvious causal model, the reinforcement is a
cause of the house being flattened.  In the contingency where the winds
do come, the house gets flattened without the reinforcement, and stays
up with it.  
\exam

Example~\ref{xam:Pandu} illustrates an even deeper problem with
Definition~\ref{actcaus}.  Note that the structural equations for
Example~\ref{xam:} are \emph{isomorphic} to the structural equations
Example~\ref{ex:ff}, provided that we interpret the variables appropriately.
Specifically, take the endogenous variables in Example~\ref{xam:Pandu} to be
$\HR$ (for ``house reinforced''), $\NW$ (for ``no wind'') and $\HU$ (for
``house up'').  Then $\HR$, $\NW$, and $\HU$ satisfy exactly the same
equations as $L$, $\ML$, and $\FF$, respectively.  That means that any
definition that just depends on the structural model is bound to give
the same answers in these two examples.  (Similar examples have been 
constructed by Hall \citeyear{Hall07}, Hiddleston
\citeyear{Hiddleston05}, and Hitchcock \citeyear{Hitchcock07}.)  This
suggests that there must 
be more to causality than just the structural equations.  
}

\xam\label{xam:bogus}
Assassin is in possession of a lethal poison, but has a last-minute
change of heart and refrains from putting it in Victim's coffee.
Bodyguard puts antidote in the coffee, which would have neutralized the
poison had there been any.  Victim drinks the coffee and survives.  
Is Bodyguard's putting in the antidote a cause of Victim surviving?
According to the preliminary HP definition, it is.  For in the contingency
where Assassin puts in the poison, Victim survives iff Bodyguard puts
in the antidote.  
\exam

Example~\ref{xam:bogus} illustrates an even deeper problem with
Definition~\ref{actcaus}.  Note that the structural equations for
Example~\ref{xam:bogus} \emph{isomorphic} to those in
Example~\ref{ex:ff}, provided that we interpret the variables appropriately.
Specifically, take the endogenous variables in Example~\ref{xam:bogus} to be
$A$ (for ``assassin does not put in poison''), $B$ (for ``bodyguard puts
in antidote''), and $\VS$ (for ``victim survives'').
Then $A$, $B$, and $\VS$ satisfy exactly the same
equations as $L$, $\ML$, and $\FF$, respectively.  That means that any
definition that just depends on the structural equations is bound to give
the same answers in these two examples.  (A similar example illustrating
the same phenomenon is given by Hall \citeyear{Hall07}.)  This suggests
that there must  
be more to causality than just the structural equations.  And, indeed, 
the final HP definition of causality allows certain contingencies to be
labeled as ``unreasonable'' or ``too far-fetched''; these contingencies
are then not considered in AC2.  

As was pointed out in \cite{Hal39}, there are problems with the HP
account.  Halpern \citeyear{Hal39} gives an alternative account, 
which is further refined and developed by Halpern and Hitchcock
\citeyear{HH11}.  I 
discuss the latter account here.  
This account builds on the assumption that the agent has, in addition to
a theory of causality 
(as modeled by the structural equations), a theory of ``normality'' or
``typicality''.  (The need to consider normality was also stressed by
Hitchcock \citeyear{Hitchcock07} and Hall \citeyear{Hall07}, and further
explored by Hitchcock and Knobe \citeyear{HK09}.)  
This theory would include statements like ``typically,
people do not put poison in coffee'' and ``typically doctors do not
treat patients to whom they are not assigned''.

As a first step to formalizing the notion of normality,
take a \emph{world} to be an assignment of values to
all the random variables.  
Thus, a world in the forest-fire example might be one where $U=(1,1,1)$,
$\ML = 1$, $L = 0$, and $\FF = 0$; the match is dropped, there is no
lightning, and no forest fire (despite the fact that either lightning or
a dropped match should be enough for there to be a forest fire).
Intuitively, a world is a complete description of a situation given the
language at our disposal (the random variables).
We want to be able to talk about one world being more normal or typical
than another.  There are many ways of doing this
(see \cite{FrH5Full} and the references therein); for
definiteness, Halpern and Hitchcock used a partial preorder $\succeq$ 
on worlds; $s \succeq s'$ means that world $s$ is at least as normal as
world $s'$.  
The fact that $\succeq$ is a partial preorder means that it is reflexive
(for all worlds $s$, we have $s \succeq s$, so $s$ is at least as normal
as itself) and transitive (if $s$ is at least as normal as $s'$ and $s'$
is at least as normal as $s''$, then $s$ is at least as normal as
$s''$).%
\footnote{If $\succeq$ were a partial order rather than just a partial
preorder, it would satisfy an 
additional assumption, \emph{antisymmetry}: $s \succeq s'$ and $s' \succeq s$
implies $s=s'$.  This is an assumption that I do \emph{not}
make here.}  A partial preorder allows to worlds to be incomparable: it may
be the case that neither $s \succeq s'$ nor $s' \succeq s$ holds.%
\footnote{This makes partial preorders more general than the \emph{ranking
functions} \cite{spohn:88} used in \cite{Hal39} to capture the relative
normality of worlds.  This extra generality turns out to be useful to
capture a number of examples.}

An {\em extended causal model\/} is
a tuple $M = (\S,\F,\succeq)$, where $(\S,\F)$ is a causal model, and
$\succeq$ is a partial preorder on worlds.  Note that
in an acyclic extended causal model, a context $\vec{u}$ determines a
world, denoted $s_{\vec{u}}$.  
We can now modify Definition~\ref{actcaus} slightly to take the relative
normality of worlds (as given by $\succeq$) into account by taking 
$\vec{X}=\vec{x}$ to be a \emph{cause of $\phi$
in an extended model $M$ and context $\vec{u}$} if $\vec{X}=\vec{x}$ is
a cause of 
$\phi$ according to Definition~\ref{actcaus}, except that in AC2(a), 
there must be a world $s$ such that $s \succeq s_{\vec{u}}$
and $s_{\vec{X} = \vec{x}', \vec{W} = \vec{w},\vec{u}} \succeq
s_{\vec{u}}$, where $s_{\vec{X} = \vec{x}', \vec{W} = \vec{w},\vec{u}}$
is the world that results by setting $\vec{X}$ to $\vec{x}'$ and $\vec{W}$
to $\vec{w}$ in context $\vec{u}$.
This modification can be viewed as a formalization of Kahneman and
Miller's \citeyear{KM86} observation that 
``an event is more likely to be undone by altering exceptional than
routine aspects of the causal chain that led to it''. In AC2(a), 
only worlds $s$ that are at least as normal as the actual
world $s_{\vec{u}}$ are considered, so indeed, it is ``exceptional
aspects'' that are being altered rather than ``routine aspects'' (since
altering routine aspects would presumably lead to a less normal world,
while altering exceptional aspects would lead to a more normal world).

This definition deals well with all the problematic examples in the
literature.   
\commentout{
It also allows us to go further: we can use normality to rank 
actual causes.  Doing so lets us explain the responses that people make
to queries regarding actual causation.
For
example, although the HP approach allows for multiple causes of an 
outcome $\phi$, people typically mention only one of them
when asked for a cause.  This can be viewed as the \emph{best} cause,
where best is judged in terms of normality.  

Say that world $s$ is a \emph{witness} for $\vec{X}=\vec{x}$ being a
cause of $\phi$  in context $\vec{u}$ if for some choice of
$\vec{Z}$, $\vec{W}$, $\vec{x}'$, and $\vec{w}$ for which AC2(a) and
AC2(b) hold, 
$s$ is the assignment of values to the endogenous variables that
results from setting $\vec{X} = \vec{x}'$ and $\vec{W} = \vec{w}$
in context $\vec{u}$. In other words, a witness $s$ is 
a world that demonstrates that AC2(a) holds.
In general, there may be many witnesses for $\vec{X}=\vec{x}$
being a cause of $\phi$.  
Say that $s$ is a \emph{best witness} for 
$\vec{X} = \vec{x}$ 
being a cause of $\phi$
if there is no other witness $s'$ such that $s' \succ s$.  (Note that
there may be more than one best witness.)  We can then grade candidate
causes according to the normality of their best witnesses.  We expect 
that someone's willingness to judge that $\vec{X} = \vec{x}$
is an actual cause of $\phi$ increases as a function of the 
normality of the best witness for $\vec{X} = \vec{x}$ in
comparison to the best witness for other candidate causes. Thus, we are
less inclined to judge that $\vec{X} = \vec{x}$  is an actual cause of
$\phi$ when there are other candidate causes of equal or higher rank.   

}
Consider Example~\ref{xam:bogus}.  
In the actual world, $A = 1$, $B = 1$, and $\VS = 1$: the assassin does
not put in poison, the bodyguard puts in the antidote, and the victim
survies.  The witness for $B = 1$ to be an actual cause of $\VS = 1$
is the world where $A = 0$, $B = 0$, and $\VS = 0$. 
If we make the assumption that $A$ typically takes the value 1 and $B$
typically  take the value $0$,\footnote{Although it may 
atypical for an assassin to poison a victim's drink, 
the action is morally wrong and unusual from the victim's perspective,
both of which would tend to make it atypical.}
we get a normality ordering in which the two worlds 
$(A = 1, B = 1, \VS = 1)$ and $(A = 0, B = 0, \VS = 0)$ are 
\emph{incomparable}.  
Since the unique witness for $B = 1$ to be an actual cause of $\VS = 1$
is incomparable with the actual world, our modified definition rules 
that $B=1$ is not an actual cause of $\VS=1$.

Now consider the first variant of Example~\ref{xam4} where
there are 100 doctors, none of whom treat Billy.  Intuitively, we want
to call Billy's doctor the cause of Billy's still being sick, since he
did not treat Billy, but we do not want to call the other 99 doctors
causes, despite the fact that they did not treat Billy either.   The way
this was dealt with in Example~\ref{xam4} was by not having variables in
the model corresponding to the other 99 doctors.  By having a theory of
normality in the model, we can deal with this issue even in a model that
includes variables for all the other doctors.
Consider an extended causal model with variables
$A_1, \ldots, A_{100}$ and $\MT_1,
\ldots, \MT_{100}$ in the causal 
model, where $A_i = 1$ if doctor $i$ is assigned to treat Billy
and $A_i = 0$ if he is not, and
$\MT_i = 1$ if doctor $i$ actually treats Billy on Monday, and
$\MT_i = 0$ if he does not.
Further assume that, typically,
no doctor is assigned to a given patient;
if doctor $i$ is not assigned to treat Billy, then typically doctor $i$
does not treat Billy;
and if doctor
$i$ \emph{is} assigned to Billy, then typically doctor $i$ treats Billy.
This can be captured by making the world
where no doctor is assigned to Billy and no doctor treats him more
normal than 
the 100 worlds where exactly one doctor is assigned to Billy, and that
doctor treats him, which in turn are more normal than the 
100 worlds where exactly one doctor is assigned to Billy and
no one treats him have rank 2, which in turn are more normal than 
the $100\times 99$ worlds where exactly one doctor is
assigned to Billy but some 
other doctor  treats him.
(The relative normality of the remaining worlds
worlds is irrelevant.)  In this extended model, in the context where
doctor $i$ 
is assigned to Billy but no one treats him, $i$ is the cause of Billy's
sickness  
(the world where $i$ treats Billy 
is more normal than the world where $i$ is assigned to Billy but no one
treats him), but no other doctor is a cause of Billy's sickness.
Moreover, in the context where $i$ is assigned to Billy and treats him,
then $i$ is the cause of Billy's recovery (for AC2(a), consider the world
where no doctor is assigned to Billy and none treat him).  

The use of normality also gives some insight into the distinction
between exogenous and endogenous variables (and the notion of condition
vs.~causality in the legal literature).  I earlier said that, typically,
we would take the presence of oxygen to be exogenous (or would not even
bother having a variable to describe it).  One way of understanding the
logic behind this is that the presence of oxygen is so much more normal
than its absence that there is no point in making it endogenous (and thus
considering changing it).  We can understand the distinction between
conditions and causes the same way.  Conditions represent values of
variables that are much more normal than all other possible values;
(potential) causes are quite often those things whose actual value is
somewhat abnormal or atypical.

Adding a representation of normality to the model has various other
advantages, as discussed by Halpern and Hitchcock \citeyear{HH11}.  In
particular, it allows us to capture instances of  the legal doctrine of
intervening causes.  In the law, it is held that one is not causally
responsible for some outcome when one's action led to that outcome only
via the intervention of a later agent's deliberate action, or some very
improbable event. For example, if Anne negligently spills gasoline, and
Bob carelessly throws a cigarette in the spill, then Anne's action is a
cause of the fire. But if Bob maliciously throws a cigarette in the gas,
then Anne is not considered a cause \cite{HH85}.\footnote{This example
is based on the facts of Watson v. Kentucky and Indiana Bridge and
Railroad \citeyear{Kentucky1910}.}  This example can be captured using an
extended causal model (see \cite{HH11} for details). 

Although the addition of normality to the causation picture gives a
great deal of added modeling power, it
raises the worry that it gives the modeler too much
flexibility. After all, the modeler can now render any claim that
$A$ is an actual cause of $B$ false, simply by choosing a normality order 
that assigns the actual world $s_{\vec{u}}$ a lower rank than any world
$s$ needed to satisfy AC2. Thus, the introduction of normality exacerbates
the problem of motivating and defending a particular choice of model.
Fortunately, the literature on the psychology of counterfactual reasoning 
and causal judgment goes some way toward enumerating the sorts of 
factors that constitute normality. (See, for example, 
\cite{Alicke92,Cushman09,CKS08,HK09,KM86,KF08,KT82,Mendel05,Roese97}.)
These include statistical frequency (things that occur more frequently
are more normal), moral judgments (actions that follow moral precepts
are more normal), and agreed-upon conventions (following a convention is
more normal that not); see \cite{HH10} for more discussion.

In particular, the law suggests a variety of principles for determining
the norms that are used in the evaluation of 
actual
causation.
In criminal law, norms are determined by direct legislation.
For example, if there are legal standards for the strength
of seat belts in an automobile, a seat belt that did not
meet this standard could be judged a cause of a traffic fatality.
By contrast, if a seat belt complied with the legal standard,
but nonetheless broke because of the extreme forces it
was subjected to during a particular accident, the fatality would
be blamed on the circumstances of the accident, rather than the
seat belt. In such a case, the manufacturers of the seat
belt would not be guilty of criminal negligence.
In contract law, compliance with the terms of a contract has
the force of a norm.
In tort law, actions are often judged against the standard of
``the reasonable person". For instance, if a bystander was
harmed when a pedestrian who was legally crossing the street
suddenly jumped out of the way of an oncoming car, the pedestrian
would not be held liable for damages to the bystander, since
he acted as the hypothetical ``reasonable person" would have done
in similar circumstances. (See, for example, \cite[pp.~142ff.]{HH85} for
discussion.) 
There are also a number of circumstances in which deliberate 
malicious acts of third parties are considered to be 
``abnormal" interventions,
and affect the assessment of causation. (See, for example,
\cite[pp.~68ff.]{HH85}.)

\section{Responsibility and blame}\label{sec:responsibility}
The HP definition of causality treats causality as an all-or-nothing
concept (as do all the other definitions of causality in the literature
that I am aware of).  While we can talk about the probability that $A$
is a cause of $B$ (by putting a probability on contexts), we cannot talk
about degree of causality.  This means that we cannot make some
distinctions that seem intuitively significant.  For example, as we
observed earlier, there seems to be a difference in the degree of
responsibility of a voter for a victory in an 11--0 election and a 6--5
election.   As Chockler and Halpern showed, one of the advantages of the
HP definition is that it provides a straightforward way of defining
refinements of the notion of causality that let us capture important
intuitions regarding degree of responsibility and blame.  The discussion
in this section closely follows that in \cite{ChocklerH03}.

\subsection{Responsibility}

The idea behind the definition of degree of responsibility is
straightforward:  If $A$ is not a cause $B$ then the degree 
of responsibility of $A$ for $B$ is 0.  If $A$ is a cause of $B$, then
the degree of responsibility of $A$
for $B$ is $1/(N+1)$, where $N$ is the minimal number of changes that
have to be made to obtain a contingency where $B$ counterfactually 
depends on $A$.  In the case of the 11--0 vote, the degree
of responsibility of any voter for the victory is $1/6$, since 5 changes
have to be made before a vote is critical.  If the vote were
1001--0, the degree of responsibility of any voter would be $1/501$.  On
the other hand, if the vote is 6--5, then the degree of responsibility
of each voter for is 1; each voter is critical.  

Thus, the degree of responsibility of $A$ for $B$ is a number between 0
and 1.  It is a refinement of the notion of causality.  It is 0 if and
only if $A$ is not a cause of $B$.  If $A$ is a cause of $B$, then the
degree of responsibility is positive.
Despite being a number between 0 and 1, as the voting examples make
clear, degree of responsibility does not act like probability at all.  

Here is the formal definition of degree of responsibility, from
\cite{ChocklerH03} (modified to be appropriate for extended causal
models rather than just causal models).

\dfn\label{def-resp}
The {\em degree of responsibility
of $\vec{X}=\vec{x}$ for $\phi$ in 
$(M,\vec{u})$\/}, denoted $\dr((M,\vec{u}), (\vec{X}=\vec{x}), \phi)$, is
$0$ if $\vec{X}=\vec{x}$ is 
not a cause of $\phi$ in $(M,\vec{u})$; it is $1/(k+1)$ if
$\vec{X}=\vec{x}$ is  a cause of $\phi$ in $(M,\vec{u})$ 
and there exists a partition $(\vec{Z},\vec{W})$ and a setting
$(\vec{x}',\vec{w})$ that determines a world at least as normal as
$s_{\vec{u}}$ for which AC2 holds
such that (a) $k$ variables in $\vec{W}$ have different values in $\vec{w}$
than they do in the context $\vec{u}$ and (b) there is no partition 
$(\vec{Z}',\vec{W}')$ and setting $(\vec{x}'',\vec{w}')$ satisfying AC2
such that only $k' < k$ variables have different values in $\vec{w}'$
than they do in $\vec{u}$.
\edfn

It should be clear that the degree of responsibility for the voting
examples is indeed 1/6 in the case of an 11--0 victory and 1 in the case
of a 6--5 victory.   It is easy to see that in the context $(1,1,1)$ in
the forest-fire example, where the lightning strikes, the arsonist drops
the match, and either one suffices for a fire, 
the lightning and the arsonist each have degree of responsibility $1/2$
for the fire
(assuming that the setting where the arsonist doesn't drop the match and
where the lightning does not strike are both allowable, which seems
reasonable). 
On the other hand, in the context $(1,1,2)$, where both
are needed for the fire, then the lightning and the arsonist each have
degree of responsibility 1.  Finally, in the Suzy-Billy example, since
Billy is not a cause, Billy has degree of responsibility 0.   Suzy's
degree of responsibility depends on which settings are allowable in the
extended causal model.  
If we take $\vec{W}$ to be $\{\BT, \BH\}$,
and keep both variables at their actual setting in the context, 
so that $\BT=1$ and 
$\BH=0$, then Suzy's throw becomes
critical; if she throws, the bottle shatters, and if she does not throw,
the bottle does not shatter (since $\BH=0$).  On the other hand, if 
the setting $(\ST=0,\BT=1,\BH=0)$ is not allowable in the extended
causal model (on the grounds that it requires Billy's throw to miss),
but the arguably more reasonable settings $(\ST=0, \BT=0,\BH=0)$ and
$(\ST=1,\BT=1,\BH=0)$ where Billy does not throw are allowable, then
Suzy's degree of responsibility is $1/2$, since we
must consider the contingency where Billy does not throw.
Thus, by using an appropriate extended causal model, 
an important intuition can be captured.

While the notion of degree of responsibility seems important, and
defining it in terms of the number of changes needed to make a
variable critical captures some reasonable intuitions, it is
admittedly a naive definition.  In some contexts it seems to come closer
to our intuitions to have the degree of responsibility of $\vec{X} =
\vec{x}$ decrease exponentially with the number of changes needed to
make $\vec{X} = \vec{x}$ critical, so that the degree of responsibility
would be, say, $1/2^k$ rather than $1/(k+1)$ if $k$ changes are needed.%
\footnote{I thank Denis Hilton for this suggestion.}
It may also be appropriate to assign weights to variables.  To
understand the intuition for this, consider a variant of the voting
example, where there are two voters, and each one controls a block of
votes: voter 1 
controls 8 votes and voter 2 controls 3 votes.  Moreover, each voter
must cast all his votes for one candidate; votes cannot be split. (This
is like voting in the U.S. Electoral College.  If we
consider a ``voter'' as representing a state in the Electoral College,
for all states besides Nebraska and Maine, vote splitting is not
allowed; the winner of the popular vote in the state gets all the state's
electoral votes.)  If both voters vote for Mr. B, then only voter 1 is
the cause of Mr. B's victory, and thus has degree of responsibility 1.
On the other hand, if vote splitting is allowed (so that voter 1 can be
represented by a random variable that takes on values 0, \ldots, 8,
while voter 2 can be represented by a random variable that takes on
values 0, \ldots, 3), then both voter 1 and voter 2 are causes of
Mr. B's victory; however, voter 1 has degree of responsibility 1 (his
vote is clearly critical), while  voter 2 has degree of responsibility
$1/2$ (since voter 2 becomes critical if, for example,  voter 1 splits
his vote 4--4).  Note that, if vote splitting is allowed, voter 2 would
continue to have degree of responsibility $1/2$ even if he controlled
only one vote and voter 1 controlled 1,000 votes.  

Some might think 
that voter 2 should have a lower degree of responsibility, which takes
into account the fact that he controls fewer votes.  We could capture
this by assigning different weights to voters (or, more precisely, to
the variables representing voters), and having the definition of degree of
responsibility use this weight (rather than implicitly weighting all
voters equally).  That is, the degree of responsibility of
$\vec{X}=\vec{x}$ for $\phi$ would be 1 over 1 + the sum of the weights
of the variables that need to be changed to make $X = x$ critical.
While this would give voter 2 lower responsibility, it 
requires a modeler to assign a weight to variables.
Moreover, it is not so clear that weighting the variables captures all
our intuitions here.  
For example, suppose that, as before, voter 1 controls 8 votes, voter 2
controls 3 votes, but now there is a third voter that controls 10 votes.
If voters 1 and 2 vote for Mr.~B and voter 3 votes for Mr.~G, many would
agree that it is reasonable to assign both voters 1 and 2 degree of
responsibility 1 for the outcome.  

An alternative that would not require any additional
information from the modeler would be to have the degree  of
responsibility of $\vec{X}=\vec{x}$ for $\phi$ depend on how many different
changes make $\vec{X}=\vec{x}$ critical.  For example, if voter 1
controls 8 votes and voter 2 controls 3, under this approach, voter 1
still has degree of responsibility 1, because all changes to voter 2's
votes make voter 1 
critical.  On the other hand, only three of the eight possible changes
to voter 1's vote (making the split 3--5, 4--4, or 5--3) make voter 2
critical.  Thus, voter 2's degree of responsibility would be $3/8$.  I
have not explored the implications of this modification of 
the definition.   While in the rest of the paper I use the
definition of degree of responsibility in Definition~\ref{def-resp}, a
variant may well be more appropriate in some applications.  
In particular, while Definition~\ref{def-resp} was argued to be useful for
\emph{model checking} (an approach to 
verifying the correctness of computer programs) \cite{CHK}, the variant
that assigns weights to variables turns out to be useful in another
model-checking context \cite{CGY08}.  Of course, legal applications may
have yet different requirements.

Interestingly, there is some evidence that people do use something like
the procedure defined here to ascribe responsibility \cite{GL10},
although more recently it has been argued that people take into account not
only the number of changes required to make a particular event critical, but
how many ways there are are to reach a situation where that event is 
critical \cite{ZGL12}.   The latter point could be captured in a
straightforward way by making a small modification to the definition of
responsibility.  Again, further experimentation is required to see
whether this is the ``right'' definition for legal applications.

\subsection{Blame}\label{sec:blame}
The definitions of both causality and
responsibility  are relative to an extended causal model and a context.
Thus, they implicitly assume that both are
given; there is no uncertainty.  Once we have a probability on contexts 
and causal models, we can talk about the probability of causality and
the expected degree of responsibility.  The latter notion is called 
\emph{(degree of) blame} by Chockler and Halpern \citeyear{ChocklerH03}.

Obviously, if we add probability to the picture, we must address the
question of where the probability is coming from.  In some cases it
might be objective; that is, it might come from a source with an
agreed-upon probability, like a coin toss.  Or we may have statistical
information that determines the probability.  Alternatively, the
probability may represent the agent's subjective beliefs.  
The definitions make sense with either interpretation.

To take a simple example of how probability can be used,
suppose that an agent is unsure as to whether the context in
the forest-fire example is (1,0,1), (1,1,1), or (1,1,2), and assigns
each of these contexts has probability $1/3$.   (For now let us not
worry about where the probability is coming from.)  With these
probabilities, the 
probability that the arsonist dropping the match
is a cause of fire is $2/3$ (since it is a cause in context $(1,1,1)$
and $(1,1,2)$, but not in context $(1,0,1)$.  The degree 
the degree of responsibility of the match is 0 in context
(1,0,1), 1 in the contexts (1,1,2) and (1,1,2), and $1/2$ in the context
(1,1,1), where either the lightning or the dropped match suffices to
cause the forest fire.  Thus, the degree of blame (i.e., expected degree
of responsibility) is $1/2$ ($= 0 \times 1/3 + 1 \times 1/3 + 1/2
\times 1/3$).

As I said earlier,
there are two significant sources of uncertainty for an
agent who is contemplating performing an action:
\begin{itemize}
\item 
what the true situation is (i.e., what value the exogenous variables
have); for example, a doctor may be 
uncertain about whether a patient has high blood pressure;
\item how the world works; for example, a doctor may be uncertain about
the side effects of a given medication.
\end{itemize}

In the HP framework, the ``true situation'' is determined by the context;
``how the world works'' is determined by the equations.
All the uncertainty about the equations can be encoded into the
context.  
For example, in modeling the forest fire, I used the context $U$ to
describe whether both the match and the lightning are needed for the
fire, or whether one of them suffices.
But at times it
is convenient to simply use two different causal models.  For example,
we can imagine a causal model for cancer where smoking causes cancer,
and another causal model where cancer is unrelated to smoking
While we now
believe that the first causal model more accurately depicts reality, at
one point there was some doubt.

Define a \emph{situation} to be a pair of the form $(M,\vec{u})$,
where $M$ is an extended causal model and $\vec{u}$ is a context.
As the discussion above suggests, an agent has uncertainty regarding the
true situation.  I thus take an agent's 
uncertainty to be modeled by a pair  $(\K,\Pr)$, where $\K$ 
is a set of situations 
and $\Pr$ is a probability distribution over $\K$.
In the forest-fire example, in all the situations, the causal model was
the same, but in general this need not be the case.  
Intuitively, $\K$ describes the situations that the agent considers
possible before $\vec{X}$ is set to $\vec{x}$.
(Note that the situation $(M_{\vec{X} \gets \vec{x}},\vec{u})$ for $(M,
\vec{u}) \in \K$  are those 
that the agent considers possible after $X$ is set to $x$.)  
The degree  of blame that setting $\vec{X}$ to $\vec{x}$ has for $\phi$ is 
then the expected degree
of responsibility of $\vec{X}=\vec{x}$ for $\phi$ in 
$(M_{\vec{X} \gets x},\vec{u})$,   
taken over the situations $(M,\vec{u}) \in \K$.

\dfn
The {\em degree of 
blame of setting $\vec{X}$ to $\vec{x}$ for $\phi$ relative to epistemic state
$(\K,\Pr)$\/}, denoted $\db(\K,\Pr,\vec{X} \gets \vec{x}, \phi)$, is
$$\sum_{(M,\vec{u}) \in \K}
\dr((M_{\vec{X} \gets \vec{x}}, \vec{u}), \vec{X} = \vec{x}, \phi)
\Pr((M,\vec{u})).$$ 
\end{definition} 
In this definition, it is perhaps best to think of $\vec{X}=\vec{x}$ as indicating
that some action has been performed.  Thus, if we are trying to decide
to what extent an agent who performs a particular action is to blame for
an outcome, we can take $X$ to be a random variable that indicates
whether the action is performed (so that $X=1$ if the action is
performed and $X=0$ otherwise) and $\phi$ to be the outcome of
interest.  To determine the degree of blame attached to $X \gets x$, we first
consider what situations the agent considers possible \emph{before} the
action is performed, and how likely each one of them is (according to
the agent).  This is given by $(\K,\Pr)$.  We then consider the agent's
degree of responsibility in each model that arises if the action is
actually performed.  If $(M,u)$ is one of the situations the agent
considers possible before performing the action, after performing the
action, the situation is described by $M_{X \gets 1}$; we thus consider
the degree of responsibility of $X=1$ for the outcome $\phi$ in the
causal model $M_{X \gets 1}$.    It may also make sense to put a
probability on the situations that arise \emph{after} the action is
performed.  I return to this issue below, after considering a few examples.

\xam Suppose that we are trying to compute the degree
of blame of Suzy's throwing the rock for the bottle shattering. 
Suppose that the only causal model that Suzy considers possible is
essentially like the second model in Example~\ref{xam2} (with $\SH$ and
$\BH$), with some minor modifications: 
$\BT$ can now take on three values, say 0, 1, 2; as before, if
$\BT = 0$ then Billy doesn't throw, if $\BT = 1$, then Billy does throw,
and if $\BT = 2$, then Billy throws extra hard.  Assume that the causal
model is such that if $\BT = 1$, then Suzy's rock will hit the bottle
first, but if $\BT =2$, they will hit simultaneously.  Thus, $\SH = 1$
if $\ST = 1$, and $\BH = 1$ if $\BT = 1$ and $\SH = 0$ or if $\BT = 2$.
Call this structural model $M$.

At time 0, Suzy considers the following four situations equally likely:
\begin{itemize}
\item $(M, \vec{u}_1)$, where $\vec{u}_1$ is such that Billy already
threw at time 0 (and hence the bottle is shattered);
\item $(M,\vec{u}_2)$, where the bottle was whole before
Suzy's throw, 
and Billy throws extra hard, so 
Billy's throw and Suzy's throw hit the bottle
simultaneously (this essentially gives the first model in
Example~\ref{xam2});
\item $(M,\vec{u}_3)$, where the bottle was whole before Suzy's throw,
and Suzy's throw hit before Billy's throw (this essentially gives the
second model in Example~\ref{xam2}); and
\item $(M,\vec{u}_4)$, where the bottle was whole before
Suzy's throw, and Billy did not throw. 
\end{itemize}
The
bottle is already shattered in $(M,\vec{u}_1)$ before Suzy's
action,
so Suzy's throw is not a cause of the bottle shattering, and her degree
of responsibility for the shattered bottle is 0.
Suzy's degree of responsibility in $(M,\vec{u}_2)$ is $1/2$---both Suzy
and Billy are equally responsible.  In $(M,\vec{u}_4)$, Suzy's
degree of responsibility is clearly 1; Billy's not throwing makes Suzy's
throw critical.  As discussed earlier, Suzy's degree of responsibility
in $(M,\vec{u}_3)$ depends on the ranking function $\kappa$.  If the
setting $(\ST=0, \BT=1, \BH=0)$ is allowable, then her degree of
responsibility  is $1$; otherwise it is $1/2$.  
In the former case, the degree of blame
is $\frac{1}{4}\cdot \frac{1}{2} +
\frac{1}{4}\cdot 1 + \frac{1}{4}\cdot 1 = \frac{5}{8}$;
in the latter case, it is
$\frac{1}{4}\cdot \frac{1}{2} + \frac{1}{4}\cdot \frac{1}{2} +
\frac{1}{4}\cdot 1 = \frac{1}{2}$.  

If instead we consider Suzy's probability on situations after the rock
is thrown and Suzy observes what happens, then she knows the outcome
with probability 1.  Thus, her degree of blame is exactly her degree of
responsibility.  If Suzy is considering whether to throw the rock and
wants to consider how much she will be to blame if the bottle shatters,
it seems more appropriate to use her prior probability on
situations.  If she is considering her degree of blame given what has
happened, then it makes more sense to use the posterior probability.
\exam

\xam\label{xam:firingsquad}
Consider a firing squad with ten  excellent marksmen.  Suppose that
marksman 1 knows that exactly one 
marksman has a live bullet in his rifle, 
and that all the marksmen will shoot.
 Thus, he considers 10
situations possible, depending on who has the bullet.  Let $p_i$ be some
marksman 1's prior probability that marksman $i$ has the live
bullet.  In situation $i$, marksman $i$ is the cause of death and has
degree of responsibility 1; in all other situations, marksman $i$ is not
the cause of death and has degree of responsibility 0.  Thus, 
the probability that marksman $i$ is the cause of death is
$p_i$, and marksman $i$'s degree of blame is also $p_i$. 
Note that if marksman 1 mistakenly believes that he has
the bullet (and thus takes $p_1=1$) when in fact it is marksman 2, then
it is possible for the degree of blame of marksman 1 (according to marksman
1) to be 1, while in fact the degree of responsibility of
marksman 1 is 0.  This shows that degree of blame is a subjective
notion, depending on an agent's subjective probability.

If the marksman never actually discovers which bullet was
live, then his prior probability is the same as the posterior
probability; it does not matter which is used to compute the degree of
blame.  If he discovers which bullet is live, then his degree of blame
will be equal to the degree of responsibility.  Finally, if he is given 
only partial information about which bullet is live, then 
an appropriate degree of blame based on his
posterior probability can be computed.
\exam


What an agent's believes is quite relevant to the law. I
briefly point out two issues here:
\begin{itemize}
\item  In some cases, we are interested not in what agents actually
  believe, but in what they should have believed (had they taken the
  trouble to get relevant information).  The definition
of blame does not change; the definition is agnostic as to what
epistemic state should be considered.  Considering the actual epistemic
state is relevant when considering intent; what the epistemic state
should have been may be more appropriate in assessing liability.
Consider, for example, a patient who dies as a result of being treated
by a doctor with a particular drug.
Assume that the patient died due to the drug's adverse side effects 
on people with high blood pressure and, for simplicity, that this was
the only cause of death. Suppose that the doctor was not 
aware of the drug's adverse side
effects.  (Formally, this means that he does not consider possible a
situation with a causal model where taking the drug causes death.)
Then, relative to the doctor's actual epistemic state, 
the doctor's degree of blame will be 0.
However, a lawyer might argue in court that the doctor should  have
known that treatment had adverse side effects for patients with high
blood pressure (because this is well documented in the literature) 
and thus should have checked the patient's blood pressure.
If the doctor had performed this test, he would have known
that the patient had high blood pressure.  With respect to the resulting
epistemic state, the doctor's degree of blame for the death is quite
high.  Of course, the lawyer's job is to convince the court that the
latter epistemic state is the appropriate one to consider when assigning
degree of blame.  

\item  Up to now I have considered the agent's prior probability.
But we can also consider the posterior probability.
The doctor's epistemic state after a patient's death
is likely to be quite different from her epistemic state before the
patient's death.  She may still consider it possible that the patient
died for reasons other than the treatment, but will consider causal
structures where the treatment was a cause of death more likely.  Thus,
the doctor will likely have higher degree of blame relative to her 
epistemic state after the treatment.
\end{itemize} 

Interestingly, all three epistemic states---the epistemic state that an
agent actually has before performing an action, the epistemic state that
the agent should have had before performing the action, and the
epistemic state after performing the action---have been considered
relevant to determining responsibility according to different legal theories
\cite[p.~482]{HH85}.   
Of course, it may not
be so easy to discover what agents know or believe.  Corporations
and defendants often try to cover-up what they knew and when they
knew it by deleting emails, not preserving documents, and so on.
Nevertheless, I think it is useful to have definitions that take
beliefs into account, and to define notions like blame that depend on
beliefs. 

Although I have only scratched the surface of these issues here, I hope
it is clear that the framework of structural equations and the
definitions based on it  provide us with some useful
tools  to address them.

\section{The NESS approach}\label{sec:NESS}

As I said, there has been extensive research on causality in both
philosophy and law.  Halpern and Pearl \citeyear{HP01b} compared
the HP approach to other work in the philosophy literature, so I focus here
on work in the legal literature.  Perhaps the best worked-out approach is
the NESS (Necessary Element of a Sufficient Set) test, originally
described by Hart and Honor\'{e}, and worked out in much greater detail
by Wright \citeyear{Wright85,wright:88,Wright01}.  Thus, I compare the
HP approach to the NESS approach here.

Wright does not provide a mathematical formalization of the NESS test; 
what I give here is my understanding of it.  
$A$ is a cause of $B$ according to the NESS test if
there exists a set $\S = \{A_1, \ldots, A_k\}$ of events,
each of which 
actually occurred, where $A = A_1$, $\Suff$ is sufficient for 
$B$, and $\Suff - \{A_1\}$ is not sufficient for $B$.  Thus, $A$ is an
element of a sufficient condition for $B$, namely $\Suff$,
and is a necessary element of that set, because any subset of
$\{A_1, \ldots, A_k\}$ that does not include $A$ is not sufficient for
$B$.%
\footnote{The NESS test
is much in the spirit of Mackie's \citeyear{mackie:65} INUS
test, according to which $A$ is a cause of $B$ if $A$ is an insufficient
but necessary part of a condition which is unnecessary but sufficient
for $B$.  However, a comparison of the two  approaches is beyond the
scope of this paper.}

The NESS test, as stated, seems intuitive and simple.
Moreover, it deals well with many examples.
Consider the forest fire.  The lightning and the arsonist are clearly
both causes; we can take the set $\Suff$ to be the singleton set consisting
of either lightning or the arsonist dropping a match.  Similarly, in 
Example~\ref{xam4}, if both Monday's doctor treating Billy and
Tuesday's doctor not treating Billy are elements of $\Suff$, then each of
them are causes. 
However, I believe that the NESS 
approach has problems with the Suzy-Billy example.  These are best
pointed out by considering a related example also considered by
Wright \citeyear{Wright85}.  This example shows that, 
although the NESS test looks quite formal, it lacks a 
definition of what it means for a set $\Suff$ of events to be
\emph{sufficient} for $B$ to occur; moreover, such a definition is sorely
needed.  

\xam\label{xam:poison} First, suppose that Pamela drinks a cup of tea
poisoned by Claire, and then dies.  It seems clear that Claire
poisoning the tea caused Pamela's death.  It seems reasonable in this
case to take $\Suff$ to consist of two events, both of which actually
occurred: 
\begin{itemize}
\item $A_1$, Claire poisoned the tea; and 
\item $A_2$, Pamela drank the tea.
\end{itemize}
Given our understanding of the world, 
it seems reasonable to say that the
$A_1$ and $A_2$ are sufficient for Pamela's death, but removing
$A_1$ results in a set that is insufficient.

But now suppose that David shoots Pamela just after she drinks the tea,
and she dies instantaneously from the shot (before the poison can take
effect).  In this case, we would want to say that David's shot is the
cause of Pamela's death, not Claire's poisoning.  Nevertheless, it would 
seem that the same argument that makes Claire's poisoning a cause without
David's shot would still make Claire's poisoning a cause even without
David's shot.  The set $\{A_1,A_2\}$ still seems sufficient for
Pamela's death, while $\{A_2\}$ is not.  

Wright \citeyear{Wright85} observes the poisoned tea would be a cause of
Pamela's death only if Pamela ``drank the tea and \emph{was alive when the
poison took effect}''.   Wright seems to be arguing that $\{A_1,A_2\}$ is in
fact \emph{not} sufficient for Pamela's death.  We need $A_3$: Pamela was
alive when the poison took effect.  While I agree that the fact that
Pamela was alive when the poison took place is critical for causality, I
do not see how it helps in the NESS test, under what seems to me the
most obvious definitions of ``sufficient''.  I would argue that
$\{A_1,A_2\}$ \emph{is} in fact just as sufficient for death as
$\{A_1, A_2, A_3\}$.  For suppose that $A_1$ and $A_2$ hold.  Either
Pamela was alive when the poison took effect, or she was not.  In the
either case, she dies.  In the former case, it is due to the poison; in
the latter case, it is not.  

But it gets worse.  While I would argue that
$\{A_1, A_2\}$ is indeed just as sufficient
for death as $\{A_1,A_2,A_3\}$, it is not clear that $\{A_1,A_2\}$ is in
fact sufficient.  Suppose, for example, that some people are
naturally immune to the poison that Claire used, and do not die from it.
Pamela is not immune.  
Then it seems that we need to add a condition $A_4$ saying that Pamela
is not immune from the poison to get a set sufficient to cause Pamela's
death.  And why should it stop there?  Suppose that the poison has an
antidote that, if administered within five minutes of the poison taking
effect, will prevent death.  Unfortunately, the antidote was not
administered to Pamela.  Do we have to add this condition to $\Suff$ to get a
sufficient set for Pamela's death?  Where does it stop?

Note that in the causal model where the only random variables are ``Claire 
poisoned the tea'', ``David shot Pamela'', ``Pamela was alive when the
poison took effect'', and ``Pamela dies'' (where the random variable has
value 1 or 0, depending on whether the event happened), with the
obvious equations, the shot is indeed a cause of death in the HP
definition, while the poisoning is not.  The argument is almost
identical to the Suzy-Billy case.  Moreover, adding additional random
variables like ``Claire is naturally immune'' or ``the antidote was
administered'' does not make a difference.   However, in general, adding
more variables might make a difference.  (Recall that, in the Suzy-Billy
example, adding the variables $\BH$ and $\SH$ was important, as is
``Pamela was alive when the poison took effect'' in this case.)  That is
why the causal model must make explicit what variables are being considered.
\exam

The issue of what counts as a sufficient cause is further complicated if
there is uncertainty about the causal structure.  

\xam\label{medication} Suppose that a doctor gives a patient a new drug
to deal with a heart ailment, and then the patient dies.  Is the new
drug the cause of death?  Even ignoring all the issues raised in
Example~\ref{xam:poison}, it  is clear that the answer depends on
whether giving the drug is a sufficient condition to cause death (given
all the other factors affecting the patient).  The NESS test seems to
implicitly assume that this is known.  For example, Wright
\citeyear{Wright06} says that the putative cause ``must be part of the
complete instantiation of the antecedent of the relevant causal law''.
But the notion of a causal law is undefined, nor is it defined when a
causal law is ``relevant''.  By way of contrast, in causal models, the
causal laws are encoded by the structural equations.
\exam

There is another (less serious) problem with the definition of NESS:
which events can go in $\Suff$.  This already arises in the
analysis of Example~\ref{xam4}.  Can $\Suff$ include ``negative'' events
like ``Tuesday's doctor did \emph{not} treat Billy''.  If so, can it
also include ``Doctor $i$ did not treat Billy'' for $i = 1, \ldots, 99$,
for the other 99 doctors who did not treat Billy?  The next example
shows that the problem is quite pervasive.

\commentout{
The next example shows that whether something is a cause may depend on 
whether $\Suff$ is allowed to contain disjunctive events.
\xam\label{xam:disjunctive} Suppose that a sensor senses the sum of the
forces applied to an object.  It reports a problem if the total force is
too high or too low.  For simplicity, suppose that the first force,
described by the random variable $F_1$, is either 0 or 1, while the
second force, is either 0, 1, or 2.  The sensor $S$ reads 1 (``good'')
if the total force $F_1 + F_2$ is either 1 or 2.  In the actual context,
$F_1 =0$ and $F_2 = 1$, so $S=1$.  According the HP model, both $F_1 =
0$ and $F_2 = 1$ are causes of $S=1$.  (To see that $F_1=1$ is a cause,
consider the contingency where $F_2 = 2$.)  Are they both causes
according to the NESS test?  Certainly $F_2 = 1$ seems to be; we can
just take $\Suff = \{F_2 = 1\}$.  What about $F_1 = 0$?  Certainly $F_1 = 0$
is not by itself sufficient for $S=1$.  But if $F_2 = 1$ is in $\Suff$, then
$F_1 = 0$ is no longer necessary.  However, if we take $\Suff$ to consist of
the disjunctive event $F_2 = 1 \lor F_2 = 2$ (which actually occurred)
and $F_1 = 0$, then this set is necessary for $S=1$, while $F_2 =1 \lor
F_2 = 2$ is not.  Thus, if disjunctive events are allowed in $\Suff$, then
$F_1 = 0$ is a cause of $S=1$.
\exam
}

\xam\label{xam:discharge} Wright \citeyear{Wright01} considers an
example where defendant 1 discharged 15 units of effluent, while defendant 2
discharged 13 units.  Suppose that 14 units of effluent are sufficient
for injury.  It seems clear that defendant 1's discharge is a cause of
injury; if he hadn't discharged any effluent, then there would have been
no injury.  What about defendant 2's discharge?  In the HP approach,
whether it is a cause depends on the random variables considered and their
possible values.  Suppose that $D_i$ is a random variable representing
defendant $i$'s discharge, for $i = 1, 2$.  If $D_1$ can only take
values 0 or 15 (i.e., if defendant 1 discharges either nothing or all 15
units), then defendant 2's discharge is not a cause.  But if $D_1$ can
take, for example, every integer value between 0 and 15, then $D_2 = 13$
is a cause (under the contingency that $D_1 = 4$, for example).

Intuitively, the decision as to whether the causal model should include
4 as a possible value of $D_1$ or have 0 and 15 as the only possible
values of $D_1$ should depend on the options available to defendant 1.
If all he can do is to press a switch that determines whether or not
there is effluent (so that pressing the switch results in $D_1$ being
15, and not pressing it result in $D_1$ being 0) then it seems
reasonable to take 0 and 15 as the only values.  On the other hand, if
the defendant can control the amount of effluent, then taking the range
of values to include every number between 0 and 15 seems more
reasonable.  

Perhaps not surprisingly, this 
issue is relevant to the NESS test as well, for the same reason.  If the
only possible values of $D_1$ are 0 or 15, then there is no set $\Suff$
including $D_2 = 13$ that is sufficient for the injury such that $D_2 =
13$ is necessary.  On the other hand, if $D_1 = 4$ is a possible event,
then there is such a set.
\exam

The second problem raised above, the question of which events can go
into $\Suff$, is easy to deal with, by simply making the set explicit.
Of course, as the examples above suggest, the choice of events will
have an impact on what counts as a cause, but that is arguably
appropriate.  Recall that causal models deal with this issue by making
explicit the signature, that is, the set of variables and their possible
values.  This gives us a set of 
primitive events of the form $X=x$.  More complicated events can be
formed as Boolean combinations of primitive events,  but it may also be
reasonable to restrict $\Suff$ to consist of only primitive events.

The first problem, defining sufficient cause, seems more serious. I
believe that a formal definition will require some of the machinery of
causal models, including structural equations.  (This point echoes
criticisms of NESS and related approaches by Pearl
\citeyear[pp. 314--315]{pearl:2k}.)  
In \cite{Hal39}, an approach to defining causality in
the spirit of Wright's definition is sketched, using the machinery of causal
models.  The approach 
delivers reasonable answers in many cases of interest and, indeed,
often agrees with the HP definition; however, I have not investigated
carefully how it does on problematic examples.%
\footnote{Interestingly, Baldwin and Neufeld \citeyear{BN03} claimed
that the NESS test could be formalized using causal models, but did not
actually show how, beyond describing some examples.  In a later paper
\cite{BN04}, they seem to 
retract the claim that the NESS test can be formalized using causal models.}

\commentout{
Fix a causal model $M$.  Recall that a primitive
event has the form $X=x$; a set of primitive events is \emph{consistent}
if it does not contain both $X=x$ and $X=x'$ for some random variable $X$ and
$x \ne x'$.  If $\Suff = \{X_1 = x_1, \ldots, X_k = x_k\}$ is a consistent
set of primitive events, then $\Suff$ is
\emph{sufficient} for $\phi$  relative to causal model $M$ if 
$M \sat [\Suff]\phi$, where $[\Suff]\phi$ is an abbreviation for
$[X_1 \gets x_1; \ldots; X_k \gets x_k] \phi$.  Roughly speaking, the
idea is to formalize the NESS test by taking $X=x$ to be a cause of
$\phi$ if there is a a set $\Suff$ including $X=x$ that is sufficient for
$\phi$, while $\Suff - \{X=x\}$ is not.  Example~\ref{xam:poison}
already shows that this will not work.  If $\CP$ is
a random variable that takes on value 1 if Claire poisoned the tea and 0
otherwise, then it is not hard to show that in the obvious causal model,
$\CP=1$ is sufficient for $\PD=1$ (Pamela dies), even if David shoots
Pamela.  To deal with this problem, we must strengthen the notion of
sufficiency to capture some of the intuitions behind AC2(b). 

Say that $\Suff$ is \emph{strongly sufficient for $\phi$ in
$(M,\vec{u})$} if  $\Suff \union \Suff'$ is sufficient for $\phi$ 
in $M$ for all 
sets $\Suff'$ consisting of primitive events $Z=z$ such that $(M,\vec{u})
\sat Z=z$.  Intuitively, $\Suff$ is strongly sufficient for $\phi$ with
in $(M,\vec{u})$ if $\Suff$ remains sufficient for $\phi$ even when
additional events, which happen to be true in $(M,\vec{u})$ are added to
it.  It may seem strange that a set $\Suff$ that is sufficient for
$\phi$ does not continue to be sufficient for $\phi$ as more events are
added to it.  But consider the Suzy-Billy example.  Given the structural
equations in that example, Billy throwing is sufficient for the bottle
shattering.  In all context where Billy throws, the bottle shatters
(assuming that the context just determines who throws and who does not).
But if $\BH=0$ is added to $\{\BT=1\}$, then the resulting set is
\emph{not} sufficient for $\BS=1$; if Billy's rock does not hit despite
Billy throwing, then the bottle does not  necessarily shatter.  Thus,
despite the fact that there is no context where Billy throws and the
bottle does not shatter, Billy throwing is not strongly sufficient for
the bottle shattering in the context where both Billy and Suzy throw
(i.e., in a context where $\BH=0$).  
We can use a similar argument to show that $\CP = 1$ is not strongly
sufficient for $\PD=1$ in the actual context, provided that the language
includes enough events.  

Recall the moral from the Suzy-Billy story.  In
order to get the ``right'' answer for causality in the presence of
preemption, there must be a variable in the language that takes on
different values depending on which of the two potential causes is the
actual cause.  In this case, we need a variable that takes on different
values depending on whether David shot; this variable 
corresponds to $\BH$ in the Suzy-Billy story.  Suppose that it
would take Pamela $t$ units of time after the poison is administered to
die; let $\AC$ be the variable that has value 1 if Pamela dies $t$ units
of time after the poison is administered and is alive before that, and
has value 0 otherwise.  Note that
$\AC=0$ if Pamela is already dead before the poison takes
effect.  In particular, if David shoots Pamela before the poison takes
effect, then $\AC=0$.  x
Then (assuming that there is no context where Pamela is immune to
the poison), although $\CP=1$ is sufficient for $\PD=1$, it is not
strongly sufficient for $\PD=1$ in the context $\vec{u}'$ where David
shoots, since $(M,\vec{u}) \sat \AC = 0$, and
$M \sat [CP\gets 1; \AC \gets 0] (\PD \ne 1)$.

The following definition is my attempt at formalizing the NESS
condition, using the ideas above.

\dfn $\vec{X} = \vec{x}$ is a {\em cause of $\phi$ in
$(M, \vec{u})$ according to the causal NESS test} if there exists a set
$\Suff$ of primitive events containing $\vec{X}=\vec{x}$ such that 
\begin{description}
\item[{\rm NT1.}] $(M,\vec{u}) \sat Y=y$ for all primitive
events $Y=y$ in $\Suff$;
\item[{\rm NT2.}] $\Suff$ is strongly sufficient for $\phi$ in
$(M,\vec{u})$;  
\item[{\rm NT3.}] $\Suff - \{\vec{X}=\vec{x}\}$ is not 
strongly
sufficient for $\phi$
in $(M,\vec{u})$.
\item[{\rm NT4.}] $\vec{X} = \vec{x}$ is minimal; no subset of $\vec{X}$
satisfies conditions NT1--3.
\end{description}
$\Suff$ is said to be a \emph{witness} for the fact that $\vec{X} =
\vec{x}$ is a cause of $\phi$ according to the causal NESS test.%
\footnote{This definition does not take into account defaults.  It can
be extended to take defaults into account by requiring that if
$\vec{u}'$ is the context showing that $\Suff - \{X=x\}$ is not strongly
sufficient for $\phi$, then $w_{\vec{u}'}$ is more normal than
$w_{\vec{u}}$.  For ease of exposition, I ignore this issue here.}
\exam

Unlike the HP definition, causes according to the causal NESS test
always consist of single conjuncts.

\pro\label{pro:singlecause} If $\{X_1 = x_1, \ldots, X_k = x_k\}$ is a
cause of $\phi$ in $M$ 
according to the causal NESS test, then $k=1$.
\epro

\prf Suppose that $\Suff$ is a witness of $\{X_1 = x_1, \ldots, X_k =
x_k\}$ being a cause of 
$\phi$ in $(M,\vec{u})$ according to the causal NESS test and, by way of
contradiction, 
that $k > 1$.  
$\Suff$ is not a witness for $\{X_1 = 1, \ldots, X_{k-1} =
x_{k-1}\}$ being a cause of $\phi$ (otherwise NT4 would be violated).
Thus, it must be the case that $\Suff' = \Suff - \{X_1 = 1, \ldots, X_{k-1} =
x_{k-1}\}$ 
is strongly sufficient for $\phi$ in $(M,\vec{u})$.  But then it follows
that that $X_k = 
x_k$ is a cause of $\phi$ in $(M,\vec{u})$ with $\Suff'$ as a witness.
To see this, note that clearly $\Suff'$ 
satisfies NT1, 
since $\Suff$ does. By assumption, $\Suff'$ is strongly sufficient for
$\phi$ in $(M,\vec{u})$, so NT2 holds.  And, also by
assumption, $\Suff' - \{X_k = x_k\} = \Suff - \{X_1 = x_1, \ldots, X_k =
x_k\}$ is not a strongly sufficient cause of $\phi$, so NT3 holds.  NT4
trivially holds.  This
shows that $\vec{X} = \vec{x}$ is not a cause of $\phi$ according to the
causal NESS test, since it does not satisfy NT4. \eprf

Since, as Example~\ref{xam3b} shows, a cause according to the HP
definition might not be a single conjunct, the HP definition and the
NESS definition are incomparable.   The requirement in the NESS
definition that there be a witness $\Suff$ such that $(M,\vec{u}') \sat
[\Suff] \phi$ in \emph{all} contexts $\vec{u}'$ is very strong.
For example, consider a vote that might be called off if the
weather is bad, where the weather is part of the context.  Thus, in a
context where the weather is bad, there is no winner, even if some votes
have been cast.  In the actual
context, the weather is fine and $A$ votes for Mr.~B, who wins the
election.  $A$ is a cause of Mr.~B's victory in this context, according
to the HP definition, but not according to the NESS test, since there is
no set $\Suff$ that includes $A$ sufficient to make Mr.~B win in all
contexts; indeed, there is no cause for Mr.~B's victory according to the
NESS test (which arguably indicates a problem with the definition).
Nevertheless, the HP definition and the NESS test (as defined above)
agree in many cases of interest; in particular, they agree on all the
examples considered in \cite{HP01b}.  The following result explains why,
in part, by giving a sufficient condition for causality according to the
NESS test to imply causality according to the HP definition.

\pro\label{pro:NESSsuff} Suppose that $X=x$ is a cause of $\phi$ in
$(M,\vec{u})$ according 
to the causal NESS test with witness $\Suff$, and there exists a
(possible empty) set $T$ of variables not mentioned in $\phi$ or $\Suff$
and a context $\vec{u}'$ such that: 
\begin{itemize}
\item[SC1.] $\Suff - \{X=x\}$ is not a sufficient condition for $\phi$
in $\vec{u}'$; that is, $(M,\vec{u}') \sat [\Suff - \{X=x\}] \neg \phi$;
\item[SC2.] the variables in $\vec{T}$ depend depend only on the context
in $\vec{u}$ and $\vec{u}'$; that is, for all $\vec{t}$, $\vec{T}'$
disjoint from $\vec{T}$, and $\vec{t}'$,
we have $(M,\vec{u}) \sat \vec{T} = \vec{t}$ iff $(M,\vec{u}) \sat
[\vec{T}' \gets \vec{t}'] (\vec{T} = \vec{t})$, and similarly for
context $\vec{u}'$;
\item[SC3.] $\phi$ is determined by $\vec{T}$ and $X$ in contexts
$\vec{u}$ and $\vec{u}'$; that is, for all $\vec{t}$, $\vec{T}'$
disjoint from $\vec{T}$ and $X$, $x'$, and $\vec{t}'$,
we have  $(M,\vec{u}') \sat [\vec{T} \gets \vec{t}, \vec{T}' \gets
\vec{t}', X = x'] \phi$ iff 
$(M,\vec{u}) \sat [\vec{T} \gets \vec{t}, \vec{T}' \gets \vec{t}', X =
x'] \phi$; 
\item[SC4.] in context $\vec{u}$, $\Suff - \{X\gets x\}$ depends 
only on $X \gets x$; that is, for all $\vec{T}'$ and $\vec{t}'$, we have 
$(M,\vec{u}) \sat [\vec{X} \gets x; \vec{T}' \gets \vec{t}']\Suff$.
\end{itemize}
Then 
$X=x$ is a cause of $\phi$ in $(M,\vec{u})$ according to the HP definition.
\epro

\prf Suppose that the hypothesis of the proposition holds.
By SC1, $(M,\vec{u}') \sat [\Suff - \{X=x\}] \neg \phi$.
Choose $x'$ such that
$(M,\vec{u}') \sat [\Suff - \{X=x\}](X=x')$.  I claim that we must have
$x\ne x'$.  For if $(M,\vec{u}') \sat [\Suff - \{X=x\}](X=x)$, then
$(M,\vec{u}') \sat [\Suff]\neg \phi$, contradicting the
assumption that $\Suff$ is strongly sufficient for $\phi$.  Let
$\vec{W}$ consist of all the variables in $\Suff$ other
than $X$, together with the set $\vec{T}$ that satisfies SC2--4;
let $\vec{Z}$ consist of all the remaining endogenous variables.
Let $\vec{w}$ be such that 
$(M,\vec{u}') \sat [\Suff - \{X=x\}](\vec{W} = \vec{w})$. 
Note that $\vec{W} = \vec{w}$ subsumes (i.e., includes all the
assignments in) $\Suff - \{X=x\}$.
It follows that $(M,\vec{u}') \sat [X \gets x', \vec{W}
\gets \vec{w}]\neg \phi$.  
By SC3, we must have  $(M,\vec{u}) \sat [X \gets x', \vec{W} \gets
\vec{w}]\neg \phi$.  Thus, AC2(a) holds.  For AC2(b), let $\vec{W}'$ be
an arbitrary subset of $\vec{W}$ and let $\vec{Z}'$ be an arbitrary
subset of $\vec{Z}$.  As in the statement of AC2(b), suppose that
$(M, \vec{u}) \sat \vec{Z} = \vec{z}^*$.  We want to show that
$(M, \vec{u}) \sat [X \gets x, \vec{W}' \gets \vec{w}, \vec{Z}' \gets
\vec{z}^*]\phi$.  Let $\vec{T}^* = \vec{T} - \vec{W}'$.  Suppose that
$(M,\vec{u}) \sat \vec{T}^* = \vec{t}^*$.  
First note that
since $\Suff$  is strongly sufficient for $\phi$ in $(M,\vec{u})$, we
must have $(M,\vec{u}') \sat [\Suff; \vec{Z}' \gets
\vec{z}^*, \vec{T}^* \gets \vec{t}^*]\phi$.
Let $\vec{W}'' = \vec{W}' \inter \vec{T}$.  By construction,
$(M,\vec{u}') \sat \vec{W}'' = \vec{w}$.  Moreover, by SC2, we have
$(M,\vec{u}') \sat [\Suff, \vec{Z}' \gets
\vec{z}^*, \vec{T}^* \gets \vec{t}^*, \vec{W}'' \gets \vec{w}]\phi$.
Note that all the variables in $\vec{W}' - \vec{W}''$ are in $\Suff -
\{X\gets x\}$, and they are assigned the same values in $\vec{W}' =
\vec{w}$ as in $\Suff$.  Thus, it follows that 
$(M,\vec{u}') \sat [\Suff, \vec{Z}' \gets
\vec{z}^*, \vec{T}^* \gets \vec{t}^*, \vec{W} \gets \vec{w}]\phi$.
By SC3, 
$(M,\vec{u}) \sat [\Suff, \vec{Z}' \gets
\vec{z}^*, \vec{T}^* \gets \vec{t}^*, \vec{W} \gets \vec{w}]\phi$.
By SC2, it follows that $(M,\vec{u}) \sat [\Suff, \vec{Z}' \gets
\vec{z}^*, \vec{W} \gets \vec{w}]\phi$. Finally, by SC4, 
it follows that $(M,\vec{u}) \sat [X \gets x, \vec{Z}' \gets
\vec{z}^*, \vec{W} \gets \vec{w}]\phi$, as desired.
\eprf

Although  the conditions in Proposition~\ref{pro:NESSsuff} are
complicated, they hold in all the examples considered in
\cite{HP01b}.  Moreover, by Proposition~\ref{pro:singlecause},
these conditions are also sufficient to guarantee that a cause according
to the HP condition is a single conjunct.

\commentout{
To make this intuition precise, note that in the structural equation
$F_X$ for an endogenous variable $X$, the value of $X$ can, in
principle, depend on the values of all the other exogenous and
endogenous variables.  However, quite often the value of $X$ depends on
the values of only a few variables.   $X$ is said to be 
independent of $Y$ if the value of $X$ does not depend on that of $Y$.

\dfn 
If $M = (\U,\V,\F)$ is a causal model and $X, y \in\U \union \V$, then 
$X$ is \emph{independent of $Y$
in $M$} 
if, for  all values $\vec{z}$ of the variables in $\U \union \V - \{X,Y\}$
and all values $y$ and $y'$ of $Y$,
we have $F_X(\vec{z},y)= F_X(\vec{z},y')$.  
$X$ is \emph{dependent} on $Y$ if it is not independent of $Y$.
If $X, Y \in \V$, then 
$X$ \emph{potentially affects} $Y$ in $M$ if
there exist $x \in \R(X)$, $y \in \R(Y)$,  some set $\Suff$ of events,
and a context $\vec{u}$ 
such that $(M,\vec{u}) \sat [\Suff]Y=y$ and   
$(M,\vec{u}) \sat [X=x,\Suff](Y \ne y)$.
A causal model $M$ is \emph{settable}
if, for every setting $\vec{y}$ of $\vec{Y}$, there exists a context
$\vec{u}$ such that $(M,\vec{u} \sat \vec{Y} = \vec{u}$.
\edfn

The model in Example~\ref{counterexample1} is not settable.  In
particular, there is no context in which $X=0$ and $Y2$, which is
exactly the setting needed for the HP definition.

\begin{theorem} If $X = x$ is a cause of $\phi$ in
$(M, \vec{u})$ according to the causal NESS test with witness $\Suff$,
$M$ is settable, and no variable in $\Suff - \{X\}$ has a potential
effect 
then $X=x$ is a cause  of $\phi$ in $(M,\vec{u})$ according to the HP
definition. 
\end{theorem}

\prf Suppose that $X=x$ is a cause of $\phi$ in
$(M, \vec{u})$ according to the causal NESS test and $M$ is settable.
Clearly AC1 and AC3 hold.  Moreover, there is a set $\Suff$ such that
$\Suff$ is strongly sufficient for $\phi$ in $(M,\vec{u})$
and $\Suff - \{X=x\}$ is not.  Since $\Suff - \{X=x\}$ is not strongly
sufficient for $\phi$, it must be the case that there exists a context
$\vec{u}'$ such that $(M,\vec{u}') \sat [\Suff - \{X=x\}]\neg\phi$. 
Let $\vec{W}$ consist of all the first-level variables in $M$ other than
those in $\Suff$.
For each variable $W \in \vec{W}$, let $w$
be its value in context $\vec{u}'$ (so that $(M,\vec{u}' \sat W=w$).
Let $x'$ be the setting of $X$ in $\vec{u}'$.
Since
$(M,\vec{u}') \sat [\Suff - \{X=x\}]\neg\phi$, we must also have
$(M,\vec{u}') \sat [\Suff - \{X=x\}, X= x', \vec{W} = \vec{w}]\neg\phi$.
Since all the variables that depend on the exogenous variables are
either in $\Suff$ or $\vec{W}$, it follows that, once these variables
are set, the context is irrelevant.  In particular,
$(M,\vec{u}) \sat [\Suff - \{X=x\}, X= x', \vec{W} = \vec{w}]\neg\phi$.
Thus, AC2(a) holds, talking $\vec{Z}$ to consist of all variables not in
$\vec{W}$.  To show that AC2(b) holds, let $\vec{W}'$ be any subset of
the variables in $\vec{W}$.  We must show that 
all subsets $\vec{W}'$ of $\vec{W}$.
By assumption, $(M,\vec{u}') \sat [\Suff]\phi$.  Hence, 
$(M,\vec{u}') \sat [\Suff, \vec{W} = \vec{w}']\phi$.  Hence, 

We can now define the class of causal models for which a NESS cause is
guaranteed to be a cause according to the HP definition as well.
}
}

\section{Conclusions}\label{sec:conc}

Perhaps the key point of the HP definition is that causality is relative
to a model.  This allows us to tailor the model appropriately.
Suppose that a drunk 18-year-old gets killed in a single-vehicle road
accident.  Many people may focus on the drunkenness as the cause of the
accident.  We all know that you shouldn't drink and drive.  But a road
engineer may want to focus on the too-sharp curve in the road, a
politician may want to focus on the fact that the law allows
18-year-olds to drink, and a psychologist may want to focus on the
youth's recent breakup with his girlfriend.\footnote{This is a 
variant of an example originally due to Hanson \citeyear{Hanson58}.}   
To the extent that causality ascriptions are meant to be guides for
future behavior---we ascribe causes so that we know what to do and not
to do next time around---it is perfectly reasonable for different
communities to focus on different aspects of a situation.  Don't drink
and drive; don't build roads with sharp curves; don't allow 18-year-olds
to drink; and don't drive after you've broken up with your girlfriend
may all be useful lessons for different communities to absorb. 
I view it as an advantage of the HP definition that it can accommodate
all these viewpoints easily, by an appropriate choice of endogenous and
exogenous variables.
(Recall that it is only endogenous
variables---the ones that can be manipulated---that can be causes.)
The fact that the HP can easily accommodate notions like responsibility
and blame is further evidence of its usefulness.  While it remains to
flesh out the case 
that these notions can be fruitfully applied to legal settings, I am
optimistic that this is indeed the case.

\paragraph{Acknowledgments:} I thank Steve Sloman for pointing out 
\cite{KM86}, and Joe Gast, Denis Hilton,  Chris Hitchcock for interesting
discussions on causality.

\bibliographystyle{chicago}
\bibliography{z,joe,refs}
\end{document}

%% file: defn.tex
\newtheorem{THEOREM}{Theorem}[section]
\newenvironment{theorem}{\begin{THEOREM} \hspace{-.85em} {\bf :} }%
                        {\end{THEOREM}}
\newtheorem{LEMMA}[THEOREM]{Lemma}
\newenvironment{lemma}{\begin{LEMMA} \hspace{-.85em} {\bf :} }%
                      {\end{LEMMA}}
\newtheorem{COROLLARY}[THEOREM]{Corollary}
\newenvironment{corollary}{\begin{COROLLARY} \hspace{-.85em} {\bf :} }%
                          {\end{COROLLARY}}
\newtheorem{PROPOSITION}[THEOREM]{Proposition}
\newenvironment{proposition}{\begin{PROPOSITION} \hspace{-.85em} {\bf :} }%
                            {\end{PROPOSITION}}
\newtheorem{DEFINITION}[THEOREM]{Definition}
\newenvironment{definition}{\begin{DEFINITION} \hspace{-.85em} {\bf :} \rm}%
                            {\end{DEFINITION}}
\newtheorem{CLAIM}[THEOREM]{Claim}
\newenvironment{claim}{\begin{CLAIM} \hspace{-.85em} {\bf :} \rm}%
                            {\end{CLAIM}}
\newtheorem{EXAMPLE}[THEOREM]{Example}
\newenvironment{example}{\begin{EXAMPLE} \hspace{-.85em} {\bf :} \rm}%
                            {\end{EXAMPLE}}
\newtheorem{REMARK}[THEOREM]{Remark}
\newenvironment{remark}{\begin{REMARK} \hspace{-.85em} {\bf :} \rm}%
                            {\end{REMARK}}

\newcommand{\thm}{\begin{theorem}}
\newcommand{\lem}{\begin{lemma}}
\newcommand{\pro}{\begin{proposition}}
\newcommand{\dfn}{\begin{definition}}
\newcommand{\rem}{\begin{remark}}
\newcommand{\xam}{\begin{example}}
\newcommand{\cor}{\begin{corollary}}
\newcommand{\prf}{\noindent{\bf Proof:} }
\newcommand{\ethm}{\end{theorem}}
\newcommand{\elem}{\end{lemma}}
\newcommand{\epro}{\end{proposition}}
\newcommand{\edfn}{\bbox\end{definition}}
\newcommand{\erem}{\bbox\end{remark}}
\newcommand{\exam}{\bbox\end{example}}
\newcommand{\ecor}{\end{corollary}}
\newcommand{\eprf}{\bbox\vspace{0.1in}}
\newcommand{\beqn}{\begin{equation}}
\newcommand{\eeqn}{\end{equation}}

\newcommand{\bbox}{\vrule height7pt width4pt depth1pt}

\newcommand{\clm}{\begin{claim}}
\newcommand{\eclm}{\end{claim}}

\newcommand{\sat}{\models}

\newcommand{\union}{\cup}
\newcommand{\inter}{\cap}

\newcommand{\FF}{{\bf F}}

\renewcommand{\phi}{\varphi}

\newcommand{\A}{{\cal A}}

\newcommand{\F}{{\cal F}}

\newcommand{\K}{{\cal K}}

\newcommand{\R}{{\cal R}}

\newcommand{\U}{{\cal U}}
\newcommand{\V}{{\cal V}}

\renewcommand{\>}{\rangle}

\newcommand{\ol}{\setlength{\itemsep}{0pt}\begin{enumerate}}
\newcommand{\eol}{\end{enumerate}\setlength{\itemsep}{-\parsep}}
\newcommand{\ul}{\setlength{\itemsep}{0pt}\begin{itemize}}
\newcommand{\dl}{\setlength{\itemsep}{0pt}\begin{description}}
\newcommand{\edl}{\end{description}\setlength{\itemsep}{-\parsep}}
\newcommand{\eul}{\end{itemize}\setlength{\itemsep}{-\parsep}}

\newcommand{\BS}{B^{\scriptscriptstyle \cS}}

\newcommand{\commentout}[1]{}

\newcommand{\bi}{\begin{itemize}}
\newcommand{\ei}{\end{itemize}}
\newcommand{\be}{\begin{enumerate}}
\newcommand{\ee}{\end{enumerate}}